\documentclass{article}

\usepackage{arxiv}

\usepackage[utf8]{inputenc} % allow utf-8 input
\usepackage[T1]{fontenc}    % use 8-bit T1 fonts
\usepackage{hyperref}       % hyperlinks
\usepackage{url}            % simple URL typesetting
\usepackage{booktabs}       % professional-quality tables
\usepackage{amsfonts}       % blackboard math symbols
\usepackage{nicefrac}       % compact symbols for 1/2, etc.
\usepackage{microtype}      % microtypography
\usepackage{lipsum}
\usepackage{graphicx}
\graphicspath{ {./images/} }
\usepackage{paracol}
\usepackage{graphicx}
\usepackage[utf8]{inputenc}
\usepackage{caption}
\usepackage{amssymb}
\usepackage{amsmath}
\usepackage{algorithm}
\usepackage[noend]{algorithmic}
\usepackage{cite}
\usepackage{float}
\usepackage{caption,subcaption}
\usepackage{lineno}

\usepackage{siunitx}
\usepackage{booktabs}
\usepackage{mathtools}

\usepackage{pifont}% http://ctan.org/pkg/pifont
\newcommand{\xmark}{\ding{55}}%
\newcommand{\td}{\mathrm{td}}%
\usepackage{url}
\DeclareMathOperator{\NE}{NE}
\DeclareMathOperator*{\argmax}{argmax}

\usepackage{xr} %-hyper}

\makeatletter
\newcommand*{\addFileDependency}[1]{% argument=file name and extension
  \typeout{(#1)}
  \@addtofilelist{#1}
  \IfFileExists{#1}{}{\typeout{No file #1.}}
}
\makeatother

\newcommand*{\myexternaldocument}[1]{%
    \externaldocument{#1}%
    \addFileDependency{#1.tex}%
    \addFileDependency{#1.aux}%
}

\myexternaldocument{supplement}

\newcommand{\ttitle}{\texttt{FastSTMF}: Efficient tropical matrix factorization algorithm for sparse data}

\newcommand{\tkeywords}{data embedding, tropical matrix factorization, sparse data, matrix completion, tropical semiring, optimization strategy}

\title{\ttitle}

\author{
Amra Omanović \\
  Faculty of Computer and Information Science\\
  University of Ljubljana\\
  Večna pot 113, 1000 Ljubljana, Slovenia\\
  \texttt{amra.omanovic@fri.uni-lj.si} \\
\And
Polona Oblak \\
  Faculty of Computer and Information Science\\
  University of Ljubljana\\
  Večna pot 113, 1000 Ljubljana, Slovenia\\
  \texttt{polona.oblak@fri.uni-lj.si} \\
\And
Tomaž Curk \\
Faculty of Computer and Information Science\\
  University of Ljubljana\\
  Večna pot 113, 1000 Ljubljana, Slovenia\\
  \texttt{tomaz.curk@fri.uni-lj.si} \\
}

\date{}

\begin{document}
\maketitle

\begin{abstract}
Matrix factorization, one of the most popular methods in machine learning,  has recently benefited from introducing non-linearity in prediction tasks using tropical semiring. The non-linearity enables a better fit to extreme values and distributions, thus discovering high-variance patterns that differ from those found by standard linear algebra. 
However, the optimization process of various tropical matrix factorization methods is slow. In our work, we propose a new method \texttt{FastSTMF} based on Sparse Tropical Matrix Factorization (\texttt{STMF}), which introduces a novel strategy for updating factor matrices that results in efficient computational performance. We evaluated the efficiency of \texttt{FastSTMF} on synthetic and real gene expression data from the TCGA database, and the results show that \texttt{FastSTMF} outperforms \texttt{STMF} in both accuracy and running time.
Compared to \texttt{NMF}, we show that \texttt{FastSTMF} performs better on some datasets and is not prone to overfitting as \texttt{NMF}.
This work sets the basis for developing other matrix factorization techniques based on many other semirings using a new proposed optimization process.
\end{abstract}


\section{Introduction}
One of the most popular and widely used machine learning and data mining methods is matrix factorization. Matrix factorization methods achieve good results in various applications such as clustering, imputation of missing data, recommender systems, image processing, and bioinformatics, \textit{e.g.}, see~\cite{xu2003document, omanovic2021sparse, koren2009matrix, haeffele2014structured, brunet2004metagenes}.
The main idea of matrix factorization is to approximate the input data matrix as a product of lower-dimensional matrices, called factor matrices. Factor matrices enable a more straightforward interpretation of the rows and columns of the input matrix.
The multiplication of factor matrices is mainly defined by applying standard linear algebra~\cite{nmf}. However, some recent approaches take advantage of the tropical semiring for matrix multiplication~\cite{tmf, cancer, omanovic2021sparse}.

% graphs
Semirings are widely used to solve graph optimization problems~\cite{gondran2008graphs}. Different semirings can describe related problems depending on the definitions of addition and multiplication. Finding the \textit{shortest path} in a graph can be characterized using the $(\min, +)$ semiring, the \textit{longest path} using the $(\max, +)$ semiring (known as \textit{tropical semiring}), while finding the \textit{minimum spanning tree} using the $(\max, \min)$ semiring.
% images/path-finding-problems.png
% semiring programming
In~\cite{belle2020semiring}, authors also introduced the semiring programming (SP) framework, which uses first-order logic structures with semiring labels to combine problems from different AI disciplines. SP contains logical theory, semiring, and solver, which simplifies the development and understanding of AI systems. Compared to constraint programming, which includes the model and solver, SP allows us to combine and integrate a wide range of specifications freely.
% gpu
One more usage of semirings is for an efficient implementation of algorithms with sparse linear algebra on GPUs using semirings~\cite{gala2022gpu}. The authors show that a sparse semiring primitive can support many critical distance measures while preserving performance and memory efficiency on the GPU.

% nn
Recent research results showed that to understand and interpret \textit{the deep neural network models}, we need to understand the relevant \textit{tropical geometry}~\cite{zhang2018tropical}. Because the matrix factorization model is a kind of a simple neural network, the first step in bringing a new paradigm in the machine learning field would be to introduce semirings in matrix factorization methods, which could be later applied to understand deep models. Standard linear algebra enables quick model optimization; nevertheless, the drawback represents the linear relations found by the model. At the same time, tropical semiring produces nonlinear models, which can explain complex relations, but the optimization is costly compared to methods using standard linear algebra~\cite{omanovic2021sparse, cancer}. The main challenge is designing a matrix factorization method based on tropical semiring while achieving fast model optimization.

In our work, we propose an efficient sparse data matrix factorization method for the prediction of missing values, called \texttt{FastSTMF}, which is available on \url{https://github.com/Ejmric/FastSTMF}. The new method is based on \textit{Sparse Tropical Matrix Factorization} (\texttt{STMF},~\cite{omanovic2021sparse}), which ensures a better fit to extreme values and distributions, thus discovering high-variance patterns. At the same time, \texttt{FastSTMF} eliminates the main drawback of a slow computational performance of its ancestor \texttt{STMF}. The proposed method achieves better results by introducing a novel strategy to update the factor matrices. We evaluate its performance on synthetic and real datasets and show that \texttt{FastSTMF} outperforms \texttt{STMF} by achieving better results with less computation. We include the widely used \texttt{NMF} method in the evaluation process, and we do not include the \texttt{Cancer} method, since it cannot perform matrix completion, \textit{i.e.}, predict missing values. In conclusion, we give a thorough discussion and present plans for future work.

\section{Related work}

The most well-studied matrix factorization method is \texttt{NMF}~\cite{nmf}, where input and factor matrices are limited to non-negative values. The factor model is additive, which simplifies the interpretation of the results. There are many different variants of \texttt{NMF} designed for a diverse set of applications~\cite{bmf, pmf, ionmf, dfmf}. These methods use standard linear algebra, where addition ($+$) and multiplication ($\times$) are standard operations.  

Standard linear algebra is limited to discovering patterns that arise from linear combinations of features. The usage of other semiring operations is expected to identify additional nonlinear patterns. Such a substitution of operations has already been successfully implemented in~\cite{cancer, capricorn, latitude,omanovic2021sparse,omanovic2020application} to perform data analysis and discover interesting patterns. In our work, we use \textit{tropical semiring}  $\mathbb{R}_{\max}=\mathbb{R}\cup \{-\infty\}$ as in~\cite{omanovic2021sparse}, where we replace the standard addition ($+$ in ${\mathbb R}$) with the maximum operator ($\max$) and the standard multiplication ($\times$ in ${\mathbb R}$) with addition ($+$). Since data are usually summarized in a matrix, we will generalize the tropical operations to $\mathbb{R}_{\max}^{m \times n}$, the set of $m\times n$ matrices with tropical entries. For each matrix $R\in \mathbb{R}_{\max}^{m \times n}$, we denote by $R_{ij}$ its entry in the $i$-th row and the $j$-th column. Tropical operations give rise to \emph{matrix addition} $A \oplus B$, where the entries of the sum of matrices $A, B \in \mathbb{R}_{\max}^{m \times n}$ are defined as
\begin{linenomath}
\begin{equation*}
    (A \oplus B)_{ij} := \max\{A_{ij}, B_{ij}\},
\end{equation*}
\end{linenomath}
$i=1,\ldots,m$, $j=1,\ldots,n$, and to the \emph{matrix multiplication}  $A\otimes B$ of matrices $A \in \mathbb{R}_{\max}^{m \times p}$, $B \in \mathbb{R}_{\max}^{p \times n}$, and its entries are defined as 
\begin{linenomath}
\begin{equation*}
    (A \otimes B)_{ij} := \max\limits_{1\leq k \leq p}\{A_{ik} + B_{kj}\},
\end{equation*}
\end{linenomath}
$i=1,\ldots,m$, $j=1,\ldots,n$.
%Add factorization problem $R=U\otimes V$.

The \emph{tropical matrix factorization} problem is defined as follows:
given an input data matrix \mbox{$R\in \mathbb{R}_{\max}^{m \times n}$} and factorization rank $r$, find \emph{coefficient factor} $U \in \mathbb{R}_{\max}^{m \times r}$ and \emph{basis factor} $V \in \mathbb{R}_{\max}^{r \times n}$ such that 
\begin{linenomath}
\begin{equation*}
    \label{eq:factorization}
    R\cong U \otimes V.
\end{equation*}
\end{linenomath}
The \texttt{STMF} method uses tropical semiring on sparse data and can better fit extreme values and distributions compared to \texttt{NMF}~\cite{omanovic2021sparse}. The two key parts of \texttt{STMF} are algorithms to update the coefficient factor $U$ (algorithm \texttt{ULF}) and the basis factor $V$ (algorithm \texttt{URF}). Each of them uses the operation $(\min, +)$, defined as
\begin{linenomath}
\begin{equation*}
    (A \otimes^{*} B)_{ij} := \min\limits_{A_{ik}, B_{kj} \text{are given}}\{A_{ik} + B_{kj}\}
\end{equation*}
\end{linenomath}
for matrices $A \in \mathbb{R}_{\max}^{m \times p}$ and $B \in \mathbb{R}_{\max}^{p \times n}$, $i=1,\ldots,m$, $j=1,\ldots,n$, which imitates multiplication with the inverse matrix, since invertible tropical matrices are rarely found. The fact that the \texttt{STMF} method computes \texttt{ULF}/\texttt{URF} for \textit{each element} of the input matrix results in low computational performance. 

Here is where we found the motivation for \texttt{FastSTMF}: the average running time of \texttt{STMF} is slow compared to \texttt{NMF}'s, see~\cite[Table 3]{omanovic2021sparse}. Note that given a factorization rank $r$, \texttt{STMF} attempts updating the factor matrices $r$ times for each element $R_{ij}$ until approximation error is decreased and therefore performs poorly in terms of running time. Our work presents several novel strategies for updating factor matrices more efficiently. %less frequently. 
We explored different approaches and the method that gave the best results, we named it \texttt{FastSTMF} (Algorithm~\ref{pseudocode_fast_stmf}).

%%%%%%%%%%%%%%%%%%%%%%%%%%%%%%%%%%%%%%%%%
\section{Methods}

The update procedure of factor matrices $U \in \mathbb{R}_{\max}^{m \times r}$ and $V \in \mathbb{R}_{\max}^{r \times n}$ to approximate a given input data matrix \mbox{$R\in \mathbb{R}_{\max}^{m \times n}$} is the crucial part of each matrix factorization algorithm. The speed of convergence depends on how effectively we make changes in the factor matrices $U$ and $V$. \texttt{STMF} updates the factor matrices $r$ times for each element $R_{ij}$ until the approximation error decreases. It starts in the upper left corner of the matrix $R$, therefore focusing the optimization more on the entries in the upper rows and leftmost columns and slowly progresses to the lower part of the matrix. Namely, for every pair of indices $(i,j)$ algorithm \texttt{ULF} (\texttt{URF}, respectively) in \texttt{STMF} contains a loop that searches for the optimal column index $k$ of the coefficient factor $U$ (row index $k$ of the basis factor $V$, respectively) to update.

\subsection{Our contribution}
We propose new update rules (defined as the \texttt{ByRow} strategy in Section~\ref{subsec:ByRow}) in which we identify for each row $i$ of $R$ only one column $j$ and the optimal $k$ to update in each iteration of $R$. We provide the triple $(i,j,k)$ as an input parameter in algorithms \texttt{F-ULF} (see~Algorithm~\ref{F-ULF}) and \texttt{F-URF} (see~Algorithm~\ref{F-URF}) to change in one update step only the entries of the $k$-th column of $U$, denoted by $U_{\cdot k}$, and the $k$-th row of $V$, denoted by $V_{k \cdot}$. This results in an update of all entries in the current approximation of the matrix $R$, based on the change of only one row of coefficient factor $U$ and only one column of basis factor $V$.

% F-ULF
\begin{algorithm}[H]
\caption{Fast update of $U, V$ based on the element from the $i$-th row of the coefficient factor $U$ (\texttt{F-ULF})}
\begin{algorithmic}
\REQUIRE data matrix $R$  $\in \mathbb{R}^{m \times n}$, coefficient factor $U$  $\in \mathbb{R}^{m \times r}$, basis factor $V$ $\in \mathbb{R}^{r \times n}$, position $(i,j,k)$
\ENSURE $U, V, f, U_{\cdot k}',  V_{k \cdot}'$ \\

\STATE $U_{\cdot k}',  V_{k \cdot}' \leftarrow U_{\cdot k}, V_{k \cdot}$

\STATE $U_{ik} \leftarrow R_{ij} - V_{kj}$ % in place change

\STATE $V_{k \cdot} \leftarrow (-U_{\cdot k})^{\top} \otimes^{*}  R$
\STATE $U_{\cdot k} \leftarrow R \otimes^{*} (-V_{k \cdot})^{\top}$

\STATE $f \leftarrow \left\lVert R - U \otimes V \right\rVert_b$
\RETURN $U, V, f, U_{\cdot k}',  V_{k \cdot}'$
\end{algorithmic}
\label{F-ULF}
\end{algorithm}

% F-URF
\begin{algorithm}[H]
\caption{Fast update of $U, V$ based on the element from the $j$-th column of the basis factor $V$ (\texttt{F-URF})}
\begin{algorithmic}
\REQUIRE data matrix $R$  $\in \mathbb{R}^{m \times n}$, coefficient factor $U$ $\in \mathbb{R}^{m \times r}$, basis factor $V$ $\in \mathbb{R}^{r \times n}$, position $(i,j,k)$
\ENSURE $U, V, f, U_{\cdot k}',  V_{k \cdot}'$ \\

\STATE $U_{\cdot k}',  V_{k \cdot}' \leftarrow U_{\cdot k}, V_{k \cdot}$

\STATE $V_{kj} \leftarrow R_{ij} - U_{ik}$

\STATE $U_{\cdot k} \leftarrow R \otimes^{*} (-V_{k \cdot})^{\top}$
\STATE $V_{k \cdot} \leftarrow (-U_{\cdot k})^{\top} \otimes^{*} R$

\STATE $f \leftarrow \left\lVert R - U \otimes V \right\rVert_b$ 
\RETURN $U, V, f, U_{\cdot k}',  V_{k \cdot}'$
\end{algorithmic}
\label{F-URF}
\end{algorithm}

The difference between \texttt{ULF}/\texttt{URF} and \texttt{F-ULF}/\texttt{{F-URF}} is in the efficiency of matrix operations and in their implementations. \texttt{ULF} and \texttt{URF} in \texttt{STMF} contain a loop that searches for the optimal $k$. In contrast, \texttt{FastSTMF} identifies $k$ in advance and provides it as an input parameter in \texttt{F-ULF} and \texttt{F-URF}, see~Algorithms~\ref{F-ULF} and~\ref{F-URF}.
\texttt{ULF}/\texttt{URF} both use whole matrices $U,V$ in all matrix operations to get new updated $U,V$.
However, we obtain the same updated $U, V$ in the proposed \texttt{F-ULF}/\texttt{F-URF} by changing only the appropriate rows/columns from $U, V$, since only specific values change in the factor matrices.
In this way, we speed up the process of computing new factor matrices by getting the same result faster (see Supplementary Figure S\ref{ulf_vs_fulf}).

\subsection{Update strategies}
\label{update_strategies}
Instead of having multiple attempts to update each element in factor matrices as in \texttt{STMF}, we propose three novel update strategies: by row (\texttt{ByRow}), by element (\texttt{ByElement}) and by entire matrix (\texttt{ByMatrix}). Modified update strategies result in more iterations through all matrix elements in the same amount of time compared to \texttt{STMF}. For each strategy, we define its own error, defined with $\rm err$, which guides the selection of factor matrices' indices to update.

\subsubsection{The \texttt{ByRow} strategy}\label{subsec:ByRow}

In the central part of our work, we investigate the different selection approaches for choosing the best indices $j$ and $k$ to update for a fixed row $i$.
In the \texttt{ByRow} strategy, we iterate through all rows of the matrix $R$. For each row $i$ of $R$, we update the factor matrices only once for each iteration over $R$.

\begin{itemize}
    \item In the \textit{sequential selection} (\texttt{SEQ}), we choose for each $i$ the smallest $j$ and $k$ that decrease the approximation error $\|.\|_b$ the most, defined as the sum of the absolute values of matrix entries (for definitions, see Subsection~\ref{subsec:metrics}). After updating the row $i$, we continue to select $(j, k)$ for the next row $i+1$.
    \item The \textit{tropical distance selection} (\texttt{TD}) aims for $(U\otimes V)_{ij}=\max\limits_{1\leq k \leq r}\{U_{ik} + V_{kj}\}$ to be as close to $R_{ij}$ in the shortest time. Therefore, when $R_{ij}$ is given, the \texttt{TD} selection chooses for each $i$ the index $j$ with the highest \textit{tropical distance}
    \begin{linenomath}
    \begin{equation*}
        \td(R,U,V,i,j) = \left| R_{ij}-\max_{1\leq k \leq r}(U_{ik}+V_{kj})\right|,
    \end{equation*}
    \end{linenomath}
    and still decreases the approximation error $\|.\|_b$. For each pair $(i,j)$, the tropical distance $\td(R,U,V,i,j)$ measures the error at position $(i,j)$ between $R$ and the current iteration $U\otimes  V$ in tropical semiring. Intuitively, we can interpret the function $\td$ as the indicator of the most suitable location in the factor matrices to update. For each pair $(i,j)$, we denote the value of index $k$ selected in $\td$ as
    \begin{linenomath}
    \begin{equation*}
    f(i,j) = \argmax_{1\leq \ell \leq r}(U_{i\ell}+V_{\ell j}).
    \end{equation*}
    \end{linenomath}
    The \texttt{TD} selection chooses $k=f(i,j)$ and when multiple $\ell$ indices give the same maximum value, the smaller $\ell$ is chosen as $k$. We expect the tropical distance selection criteria to find the tropical patterns in the matrix and correct the largest entry-wise error.
    
    \item We investigate two subvariants of \texttt{TD}, denoted
    \texttt{TD\_A} and \texttt{TD\_B}. Instead of comparing the error only locally at position $(i,j)$ as in \texttt{TD}, these subvariants will act more globally and will scan a whole row and column to find the best index $k$. We define the tropical distance for row $i$ as
    \begin{linenomath}
    \begin{equation*}
        \td_{row}(R,U,V,i) = \sum_{t=1}^{n} \td(R,U,V,i,t)
    \end{equation*}
    \end{linenomath}
    and for column $j$ as
    \begin{linenomath}
    \begin{equation*}
        \td_{col}(R,U,V,j) = \sum_{t=1}^{m} \td(R,U,V,t,j).
    \end{equation*}
    \end{linenomath}
    %This brings us to two subvariants of \texttt{TD}, denoted as \texttt{TD\_A} and \texttt{TD\_B}. 
    Both variants select the index $j$ to be the one that results in the maximum value of
    \begin{linenomath}
    \begin{equation*}
    \label{err_byrow}
    {\rm err}(R,U,V,i,j) = \td_{col}(R,U,V,j).
    \end{equation*}
    \end{linenomath}
    Since the row $i$ is predetermined and the value of $\td_{row}(R,U,V,i)$ is fixed, we are only interested in selecting the column index $j$ with the highest ${\rm err}$. 
    The variant \texttt{TD\_A} selects $k$ as the index that occurs the most times in $\td_{row}(R,U,V,i)$ and $\td_{col}(R,U,V,j)$ together. Formally,
    \begin{linenomath}
    \begin{equation*}
    k= \argmax_\ell\left({\rm count_\ell}\left(\{\!\!\{f(i,t)\colon t=1,\ldots,n\}\!\!\} \cup \{\!\!\{f(t,j)\colon t=1,\ldots,m\}\!\!\}\right)\right), 
    \end{equation*}
    \end{linenomath}
    where the function $\rm count_\ell$ returns the number of occurrences of $\ell$ in the multiset $\{\!\!\{.\}\!\!\}$.
    
     For variant \texttt{TD\_B}, we first define $j_0=\argmax_{1\leq t \leq n} ({\rm err}(R,U,V,i,t))$ and $i_0=\argmax_{1\leq t \leq m} ({\rm err}(R,U,V,t,j))$ as indices, where the maximal ${\rm err}$ is achieved in the $i$-th row and the $j$-th column. We choose $k$ as
    \begin{linenomath}
    \begin{equation*}
    k = \begin{cases}
       f(i,j_0), & (U\otimes V)_{ij_0}\geq (U\otimes V)_{i_0j},\\
       f(i_0,j), & \text{otherwise}.\\
    \end{cases}
    \end{equation*}
    \end{linenomath}
\end{itemize} 

\subsubsection{The \texttt{ByElement} strategy}
The \texttt{ByElement} strategy updates the factor matrices, when  $R_{ij}$ is given, only once by choosing the index $k=k_{ij}$ using \texttt{TD} selection.
For a fixed $(i, j)$, we compute $\td(R, U, V, i,j)$, which represents the $\text{err}$:
\begin{linenomath}
\begin{equation*}
    {\rm err}(R,U,V,i,j) = \td(R,U,V,i,j).
\end{equation*}
\end{linenomath}
In case $\rm err$ equals $0$, we have a perfect approximation of $(i,j)$ and we skip updating that element.

\subsubsection{The \texttt{ByMatrix} strategy}
On the contrary, the \texttt{ByMatrix} strategy selects only one $i$ and $j$ over the entire data matrix $R$ in each iteration through all given elements of the matrix. Namely, we define the error of every pair $(i,j)$ as
\begin{linenomath}
\begin{equation*}
\label{err_bymatrix}
{\rm err}(R,U,V,i,j) = \td_{row}(R,U,V,i) + \td_{col}(R,U,V,j) - \td(R,U,V,i,j).
\end{equation*}
\end{linenomath}
Since $(i, j)$ appears in $\td_{row}(R,U,V,i)$ and in $\td_{col}(R,U,V,j)$, we subtract $\td(R,U,V,i,j)$ in ${\rm err}$ to avoid double counting. % in~\eqref{err_bymatrix}. 
We sort all $\text{err}$ in decreasing order and select a pair of indices $(i,j)$ with the highest ${\rm err}$, while the index $k$ is chosen using \texttt{TD} selection for $(i, j)$.  Since \texttt{ByMatrix} strategy compares the errors of all matrix elements before the update, it is extremely slow. Our aim is to create a fast tropical method, thus we randomly apply \texttt{F-ULF} or \texttt{F-URF} to the chosen triple $(i,j,k)$ instead of testing them both. We halve the running time at the expense of skipping potential good indices to update.

\subsection{Comparison of strategies} 
In each of these three strategies, we choose the indices $(i, j, k)$ with the highest value $\rm err$ that decreases the approximation error and $k$ as described above.
The reason for a more time-efficient updating of the factor matrices is the introduction of $\td$-based heuristics in a given running time. %together with identifying which $(i, j, k)$ results in maximal $\rm err$.
If the selected indices $(i, j, k)$ do not decrease the approximation error, we use the indices computed from the next highest $\rm err$. 

For some of the proposed methods, we first choose to perform permuting of rows (\texttt{PermR}) or columns (\texttt{PermC}) in the data matrix.  All permutations sort rows or columns in increasing order by their minimum values, consistent with the proven efficient criteria used in \texttt{STMF} (see~\cite[Fig.~3]{omanovic2021sparse}).
In the \texttt{ByMatrix} strategy, the permutation of rows or columns is unnecessary, since the values of $\text{err}$ are always computed for all entries of the matrix, and the permuting procedure would not influence the algorithm's results.

In total, we will compare and evaluate twelve methods with various combinations of update strategies and data permutations on synthetic data. Among ten \texttt{ByRow} strategies, we chose the one that performed best, see Subsection~\ref{subsection_seq_vs_td}. We selected the preferred data matrix shape for \texttt{ByElement}, \texttt{ByRow} and \texttt{ByMatrix}, see Subsection~\ref{subsection_matrix_shape}. We summarize three final methods that gave the best results in Table~\ref{table_stmf_versions} and give their pseudocodes in Algorithm~\ref{pseudocode_fast_stmf} and in the Supplement, Section~\ref{pseudocode}. Based on additional evaluation on synthetic data we chose the best method as \texttt{FastSTMF}.

\begin{table}[htb]
    \centering
    \resizebox{\textwidth}{!}{
    \begin{tabular}{c|c|c|c|c|c}
        & \textbf{\shortstack[c]{update strategy}} & \textbf{\shortstack[c]{permutation\\ by columns}} & \textbf{\shortstack[c]{permutation\\ by rows}} & \textbf{\shortstack[c]{candidate\\ selection}}  & \textbf{\shortstack[c]{preferred\\ shape}}\\
        \hline
        \texttt{STMF} & $r$ times per element & incr. by minimum & \xmark & \xmark & \xmark \\
        \texttt{STMF\_ByElement\_PermC\_TD\_W} & once per element & incr. by minimum & \xmark  & TD & wide\\
        \texttt{STMF\_ByRow\_RandPermR\_TD\_A\_W} (\texttt{FastSTMF}) & once per row & \xmark & random  & TD\_A & wide\\
        \texttt{STMF\_ByMatrix\_NoPerm\_TD\_W}& once per matrix & \xmark & \xmark & TD & wide\\
        \hline
    \end{tabular}}
    \caption{Comparison of proposed methods with the \texttt{STMF} method given the update strategy, performed data matrix permutations, performed candidate selection %(see Subsection~\ref{subsection_seq_vs_td}) 
    and preferred data matrix shape (see Section~\ref{sec:results}).
    }
    \label{table_stmf_versions}
\end{table}

We use the Random Acol initialization strategy for the matrix $U$ as used in \texttt{STMF}~\cite{omanovic2021sparse}. Random Acol is the element-wise average strategy of randomly selected columns from the input data matrix $R$.
The convergence of the proposed algorithm \texttt{FastSTMF}, defined in Algorithm~\ref{pseudocode_fast_stmf}, is checked similarly to that of \texttt{STMF}~\cite{omanovic2021sparse}. The update rules of \texttt{FastSTMF} step-by-step reduce the approximation error, since the factor matrices $U$ and $V$ are only updated in the case when the $b$-norm monotonously decreases. Convergence is achieved when the change in error is less than a given threshold $\epsilon$.

%%%%%%%%%%%%%%%%%%%%%%%%%%%%%%%%%%%%%%%%%%%%%%%%%%%%%%%%%%%%%%%%%%
\begin{algorithm}[H]
\caption{\texttt{FastSTMF} (\texttt{STMF\_ByRow\_RandPerm\_TD\_A\_W})}
\label{pseudocode_fast_stmf}
\begin{algorithmic}
\REQUIRE data matrix $R$ $\in \mathbb{R}^{m \times n}$, factorization rank $r$
\ENSURE factorization $U$ $\in \mathbb{R}^{m \times r}$, $V$ $\in \mathbb{R}^{r \times n}$

\STATE \textbf{if} $R$ not wide \textbf{then}  transpose $R$
\STATE $perm \gets$ random permutation of indices $1 \dots m$
\STATE $R \gets R[perm, :]$
\STATE initialize $U$ and compute $V$ 
        
\WHILE {not converged}
\STATE \textbf{for} each \textit{row} $i$ of $R$
\begin{ALC@g}
\STATE $row\_indices \gets \{\!\!\{ f(i,t)\colon t=1,\ldots,n \}\!\!\}$
\STATE \textit{errors}, $columns\_indices \gets [\,], [\, ]$
\STATE \textbf{for} each \textit{column} $j$ of $R$

\begin{ALC@g}
\STATE $e \gets \td_{col}(R, U, V, j)$
\STATE \textbf{append} $e$ \textbf{to} \textit{errors}
\STATE $col\_indices \gets \{\!\!\{ f(t,j)\colon t=1,\ldots,m \}\!\!\}$
\STATE \textbf{append} $col\_indices$ \textbf{to} $columns\_indices$
\end{ALC@g}
\STATE \textbf{for} each $j$ \textbf{in} argsort(\textit{errors}) in decreasing order

\begin{ALC@g}
\STATE $k\gets \argmax_\ell\left({\rm count_\ell}\left(row\_indices \text{ }\cup \text{ } columns\_indices[j]\right)\right)$
\STATE $(U, V, f, U_{\cdot k}',  V_{k \cdot}') \gets \text{F-ULF}(R, U, V, i, j, k)$

\STATE \textbf{if} $f$ decreases \textbf{then} \textbf{break}
\STATE \textbf{else} $(U_{\cdot k}, V_{k \cdot}) \gets (U_{\cdot k}', V_{k \cdot}')$

\end{ALC@g}
\begin{ALC@g}
\STATE $(U, V, f, U_{\cdot k}',  V_{k \cdot}') \gets \text{F-URF}(R, U, V, i, j, k)$
\end{ALC@g}

\begin{ALC@g}
\STATE \textbf{if} $f$ decreases \textbf{then} \textbf{break}
\STATE \textbf{else} $(U_{\cdot k}, V_{k \cdot}) \gets (U_{\cdot k}', V_{k \cdot}')$
\end{ALC@g}

\end{ALC@g}
\ENDWHILE

\STATE \textbf{if} $R$ transposed \textbf{then} $(U, V) \gets (V^{T}, U[perm^{-1}, :]^{T})$
\STATE \textbf{else} $(U, V) \gets (U[perm^{-1}, :], V)$
%\RETURN $U[perm^{-1}, :], V$
\RETURN $U, V$
\end{algorithmic}
\end{algorithm}

\subsection{Synthetic data}
We create two groups of synthetic datasets $R_\lambda$ with matrices of rank three: smaller, of size $200 \times 200$, and larger, of size $10 000 \times 300$. In each group, the dataset $R_\lambda$ is computed using the following formula: 
\begin{linenomath}
\begin{equation*}
R_\lambda = \lambda (A \otimes B) + (1 - \lambda) (A \cdot B),
\end{equation*}
\end{linenomath}
where $A$ and $B$ are two random non-negative matrices sampled from a uniform distribution over $[0, 1)$, and $A \otimes B$ is the tropical product and $A \cdot B$ is the standard product of generated matrices. Thus, when $\lambda=0$ the dataset $R_\lambda$ has a completely \textit{linear structure} and $\lambda=1$ represents a completely \textit{tropical structure} of $R_\lambda$. Note that since the entries of $A$ and $B$ are uniformly distributed in $[0,1)$, the average entry of $A\otimes B$ will be in general larger than the average entry of $A\cdot B$. Since we are interested in tropical factorization, we also consider $\lambda=0.5$ to generate data matrix $R$ with a more tropical than linear structure, called \textit{mixture}. 
We generate $50$ small synthetic datasets and $100$ large synthetic datasets. 
We split the large synthetic data in $50$ matrices representing the training set and $50$ representing the validation set.

\subsection{Real data}
In our work, we use the following preprocessed gene expression datasets from The Cancer Genome Atlas (TCGA) database~\cite{rappoport2018multi}: Acute Myeloid Leukemia (AML), Breast Invasive Carcinoma (BIC), Colon Adenocarcinoma (COLON), Glioblastoma Multiforme (GBM), Liver Hepatocellular Carcinoma (LIHC), \linebreak Lung Squamous Cell Carci\-no\-ma (LUSC), Ovarian Serous Cystadenocarcinoma (OV), Skim Cutaneous Melanoma (SKCM) and Sarcoma (SARC).

Similarly as in the evaluation of \texttt{STMF}, we use feature agglomeration~\cite{murtagh2014ward} to merge similar genes and to minimize the influence of non-informative genes. In addition, we compare the performance of methods when different number of features are given. We merge similar genes into 100 features, representing our \textit{small} real data matrices, and merge into 1000 features for the \textit{large} real data matrices. We use the 50 known PAM genes~\cite{parker2009supervised} representing our features for the BIC data. All the real data is in the form of $patients \times meta\text{-}genes$. 

%%%%%%%%%%%%%% METRICS
\subsection{Evaluation metrics}\label{subsec:metrics}

%{b-norm}
\textbf{$b$-norm} is defined as a sum of the absolute values of matrix entries
$||W||_b=\sum_{i,j}\vert W_{ij} \vert$, and it can be applied as an objective function in matrix factorization methods. As in~\cite{omanovic2021sparse}, we use the $b$-norm to minimize the approximation error of \texttt{FastSTMF}.

\textbf{Normalized error ($\NE$).} For each of the tested method, we measure \textit{normalized error} $\NE_t$ compared to baseline \texttt{STMF}'s error at timestamp $t$ as:
\begin{linenomath}
\begin{equation*}
    \NE_t = \frac{e_{t} - \gamma_{t_{\max}}}{\gamma_{t_{\text{init}}} - \gamma_{t_{\max}}},
\end{equation*}
\end{linenomath}
where $\gamma_t$ is the approximation error of \texttt{STMF} at the timestamp $t$ and $e_{t}$ is the error of a specific method at the timestamp $t$ computed using the $b$-norm. We choose $t_{\text{init}}$ as the time after initialization in \texttt{STMF}.
$\NE_t$ allows us to compare different methods on multiple datasets and observe how fast the error drops below $0$, which is $\NE_{t_{\max}}$ for \texttt{STMF}.
Negative $\NE$ values imply that the specific method achieves better results in less time compared to \texttt{STMF}.

\textbf{Distance correlation (DC)} is a well-known measure for detecting a wide range of relationships, including nonlinear~\cite{szekely2009brownian}. The DC  coefficient is defined on the interval [0, 1] and equals 0 when the variables are independent. The coefficient tells us the strength of association between the original and approximated matrix. In our previous study~\cite{omanovic2021sparse}, we used DC to evaluate the predictive performance of different matrix factorization methods.

\textbf{Root-mean-square error (RMSE)} measure is used to assess the approximation error of the training data, denoted RMSE-A, and to rate the prediction error on the real test data, denoted RMSE-P. The value of RMSE is always non-negative, and 0 indicates a perfect fit to data. It is a commonly used metric to evaluate matrix factorization methods~\cite{bokde2015matrix}.

\textbf{Critical difference (CD)} graph defined by Demšar~\cite{demvsar2006statistical} is used to compare the performance of multiple algorithms on multiple datasets. CD is computed using a statistical test for the average performance ranks of methods over different datasets for the significance level $\alpha = 0.05$. We rank the methods by the time needed to achieve \texttt{STMF}'s final $\NE_{t_{\max}}$ and by their final $\NE_{t_{\max}}$. When observing the CD graph, the group of methods not connected to another group indicates that there is a significant difference between the two groups. It is a rigorous method for evaluating algorithms since it only observes rankings in performance without considering how much one method is numerically better than others.

\textbf{Bootstrap} is a resampling technique used to compute confidence intervals of statistics such as mean, standard deviation, or other. It is performed by sampling a dataset with replacement without assumptions about distributions or variances~\cite{hesterberg2011bootstrap}. In our work, we use bootstrap to estimate the confidence interval of the mean statistic.

\subsection{Evaluation}
We compare all our proposed methods with \texttt{STMF} on synthetic datasets, while on real datasets we compare the proposed \texttt{FastSTMF} with \texttt{NMF} and \texttt{STMF}.
Since we cannot directly compare the tested methods and \texttt{STMF} by the number of iterations, we measure their errors $\NE_t$ at regular wall clock time points $t$ during execution. We limit the execution to the wall clock running time ($t_{\max}$). We then observe the running time needed to achieve \texttt{STMF}'s final $\NE_{t_{\max}}$ and use the running time to rank the methods. We also rank the methods by their final $\NE_{t_{\max}}$ error at running time $t_{\max}$.
For large synthetic datasets and large real datasets, we set the maximum running time to $t_{\max}=600 s$. For small synthetic datasets, we set the time to $t_{\max}=100 s$. For small real datasets, we set the time to $t_{\max}=300 s$, since all methods converge faster on smaller data. We evaluate the proposed methods on a server with 48 cores on two 2.3 GHz Intel CPUs and 492 GB RAM. For a fair comparison and to prevent the methods from interrupting each other, we run experiments in parallel and ensure that the CPUs never reaches full load.

We perform ten runs with different initializations of factor matrices on each dataset and select the median $\NE$. Experiments on synthetic data used factorization rank three, but the experiments on real data were performed for the selected optimal factorization rank, already chosen in~\cite[see Table~3]{omanovic2021sparse}. We randomly and uniformly mask $20\%$ of data as missing for all datasets, which we use as a test set for the prediction task.

\section{Results}
\label{sec:results}
We use synthetic data to improve all proposed algorithms and to choose which method to name \texttt{FastSTMF}. We then performed experiments on small and large real data matrices to compare the performance of the models obtained by \texttt{STMF}, \texttt{FastSTMF}, and \texttt{NMF}. We do not compare with the \texttt{Cancer} method, as it cannot predict missing values.

\subsection{Synthetic data}
First, we demonstrate the benefits of using tropical distance versus sequential selection of indices on small synthetic matrices.
Second, we use large synthetic training data to determine the data matrix shape (wide or tall matrix) where each proposed method achieves the smallest error (we refer to it as the method's~\emph{preferred shape}) of the input data matrix. We then include the appropriate transposition of the input data at the beginning of each method and evaluate the performance on the synthetic validation data.

\subsubsection{Comparing selections in the \texttt{ByRow} strategy}
\label{subsection_seq_vs_td}

\begin{figure}[htp]
\centering
\includegraphics[]{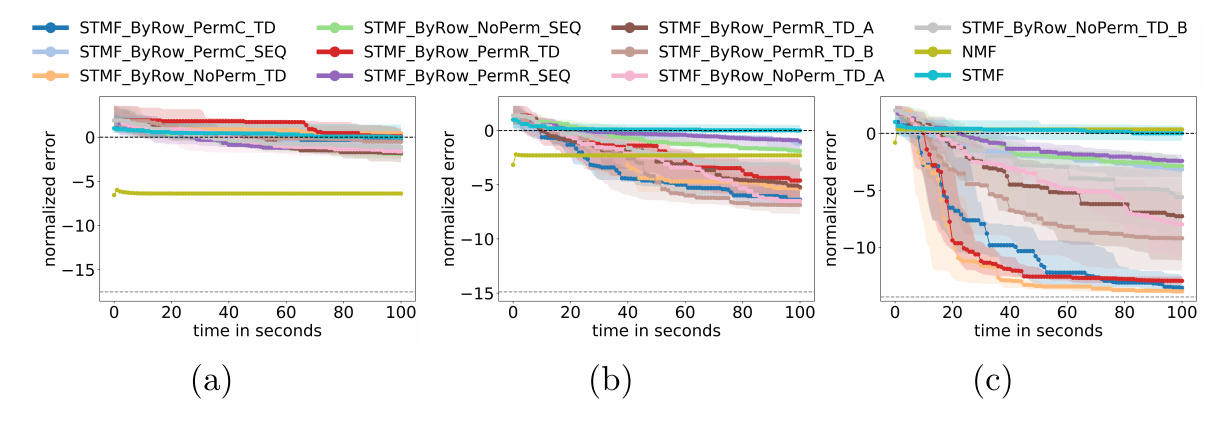}
\caption{Comparison of the median $\NE$ and the first and third quantiles between sequential selection (SEQ) of indices, tropical distance (TD) selections, and \texttt{NMF} for different \texttt{ByRow} strategies on \textit{one} small synthetic dataset for: (a) $\lambda = 0$, linear structure, (b) $\lambda = 0.5$, mixture and (c) $\lambda = 1$, tropical structure.
The \textit{black} horizontal dashed line represents \texttt{STMF}'s final $\NE_{t_{\max}}$, while the \textit{grey} horizontal dashed line represents perfect approximation, \textit{i.e.}, approximation error being equal to 0.}
\label{NE_lambda_0_1}
\end{figure}

We compare selections \texttt{TD}, \texttt{TD\_A}, \texttt{TD\_B} and \texttt{SEQ} as described in Subsection~\ref{subsec:ByRow}.
The performance between tropical distance selection, sequential selection and \texttt{NMF} differs depending on different $\lambda$ values (see Figure~\ref{NE_lambda_0_1}). 
As expected, when $\lambda=0$, \texttt{NMF} achieves best approximation since the data has linear structure, while for tropical structure ($\lambda=1$), \texttt{TD} methods achieve almost perfect approximation. In the case of mixed linear-tropical structure ($\lambda=0.5$), \texttt{TD\_A} and \texttt{TD\_B} selections have best results.
\texttt{NMF} uses the Frobenius norm in the optimization process, which is monotone decreasing. The reason why the \texttt{NMF}'s error slightly increases in Figure~\ref{NE_lambda_0_1} is because we are using the $b$-norm in $\NE$ computation, which allows us to compare all methods fairly (see Supplementary Figure S\ref{nmf_bnorm_vs_fro}).

\begin{figure}[htp]
\centering
\includegraphics[]{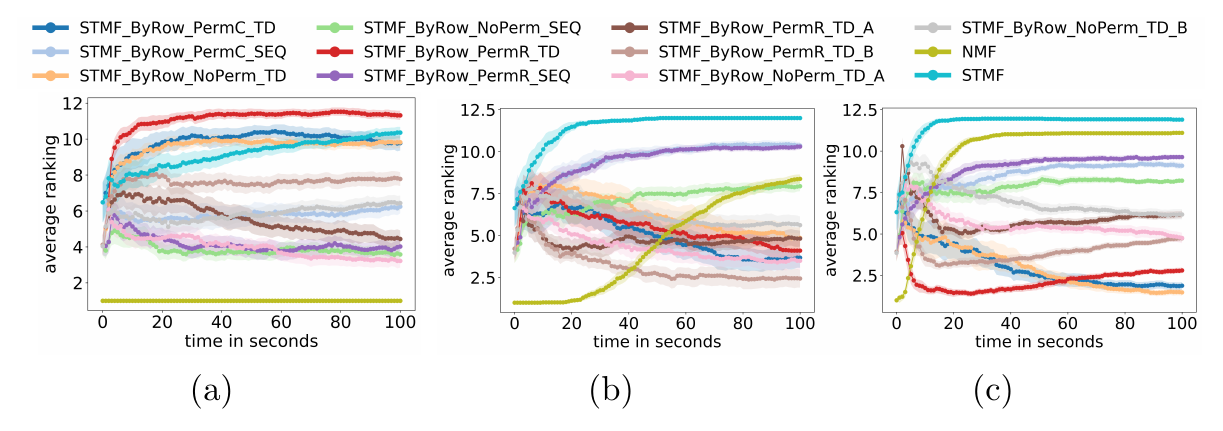}
\caption{Average rankings with $95\%$ confidence interval from bootstrap on 50 datasets for: (a) $\lambda = 0$, linear structure, (b) $\lambda = 0.5$, mixture, and (c) $\lambda = 1$, tropical structure.}
\label{rankings_lambda_0_1}
\end{figure}

The results of $\NE$ for all 50 datasets are summarized in a graph showing the ranking of methods (Figure~\ref{rankings_lambda_0_1}). For each second of running time, we rank the methods by their current $\NE$ and compute the average ranking and the $95\%$ confidence interval.

\begin{figure}[H]
    \centering
    \includegraphics{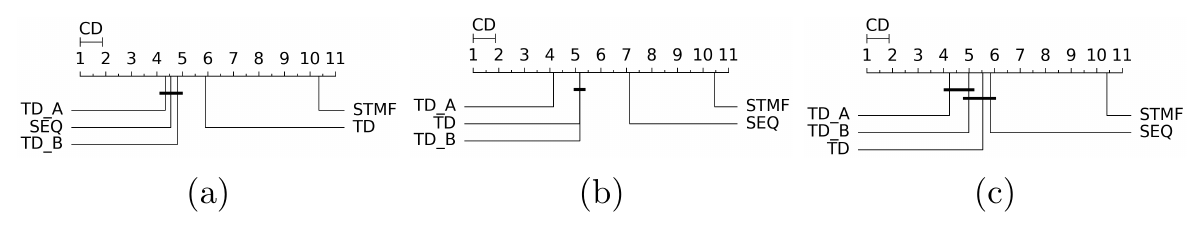}
    \caption{CD graph of averaged rankings for all three values of  $\lambda\in \{0, 0.5, 1\}$ on 50 datasets. Ranking of methods using different criteria: (a) time needed to achieve \texttt{STMF}'s final $\NE_{t_{\max}}$, (b) final $\NE_{t_{\max}}$ of its method and (c) averaged rankings of both (a) and (b).}
    \label{CD_lambda_0_1}
\end{figure}

We rank the methods by the time needed to achieve \texttt{STMF}'s final $\NE_{t_{\max}}$ and by their final $\NE_{t_{\max}}$. If value $\lambda$ is known in advance, we can choose the specific selection method as the best choice, as shown in Supplementary Figures S\ref{supp_CD_lambda_0}, S\ref{supp_CD_lambda_0.5}, and S\ref{supp_CD_lambda_1}. 
We show in Figure~\ref{CD_lambda_0_1} the average ranking for all $\lambda$ values that may be used when there is no prior knowledge about $\lambda$. Figure~\ref{CD_lambda_0_1}(c) shows the average ranking when both criteria \texttt{STMF}'s final $\NE_{t_{\max}}$ and final $\NE_{t_{\max}}$ are taken into account. Methods that use the \texttt{TD\_A} selection achieve the best results according to all three criteria. %, see CD graphs in Figure~\ref{CD_lambda_0_1}.
Based on this, we focus on the \texttt{TD\_A} methods in the \texttt{ByRow} strategies. 

The \texttt{TD\_A} selection contains two methods: \texttt{STMF\_ByRow\_PermR\_TD\_A} and \linebreak  \texttt{STMF\_ByRow\_NoPerm\_TD\_A}, which work best on the mixed linear-tropical structure ($\lambda=0.5$).
The \texttt{NoPerm} version of the method achieves better rankings than \texttt{PermR} for all three values of $\lambda$ (see Figure~\ref{rankings_lambda_0_1}). Thus, we continue our synthetic experiments with the $\lambda=0.5$ and the \texttt{NoPerm} version.

Since experiments are performed on randomly generated synthetic datasets where the order of rows bear no meaning, we notice that the \texttt{NoPerm} is a special case of \texttt{RandPerm}.
On the contrary, in real datasets, rows can have meaning and their initial permutation (\texttt{NoPerm}) can contain a structure that could lead to a biased approximation. An example of such a real-world dataset would be the gene expression data matrix, where cells are sorted by their class (\textit{e.g.}, healthy cells first and then cancer cells). In limited running time, it could happen that the \texttt{ByRow} strategy would fit only a single class and not the entire matrix. We decide to address this issue by performing a random permutation of the rows \texttt{RandPerm} before fitting the data.
The motivation comes from other machine learning resampling procedures, such as cross-validation, where data are shuffled randomly at the beginning, or in stochastic gradient optimization where data can be shuffled after each epoch to reduce overfitting and variance~\cite{bottou2012stochastic}.
Based on this, we decide to use \texttt{STMF\_ByRow\_RandPerm\_TD\_A} as the best representative of the \texttt{ByRow} strategy in the rest of the paper.

\subsubsection{Determining the preferred shape on training data}
\label{subsection_matrix_shape}
\begin{figure}[H]
    \centering
    \includegraphics{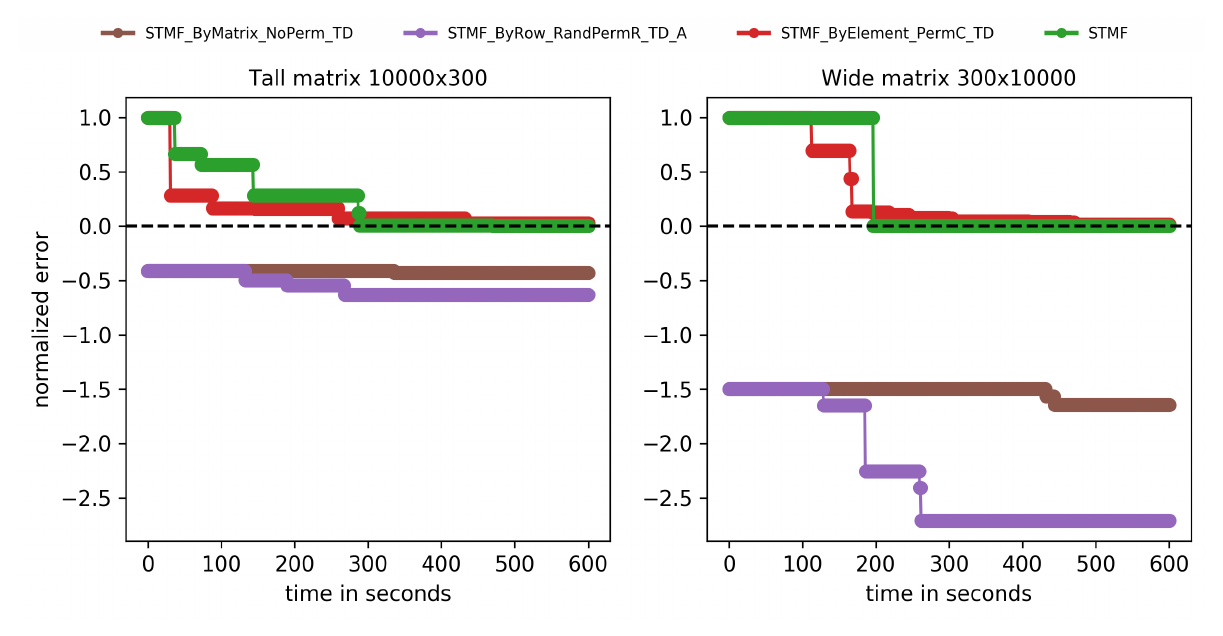}
    \caption{Comparison of median $\NE$ among different methods on one example of large synthetic rank three training data matrix presented in a \textit{tall} matrix of size $10 000 \times 300$ and a \textit{wide} matrix of size $300 \times 10 000$.}
    \label{NE_tall_vs_wide}
\end{figure}

\begin{figure}[H]
    \centering
    \includegraphics{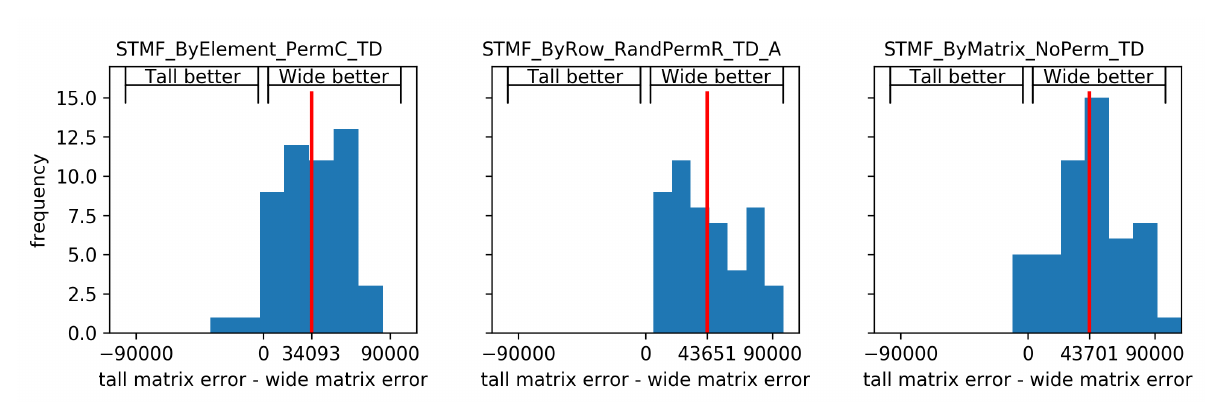}
    \caption{Distribution of differences between final errors on tall and wide matrices, where negative values mean that error on \textit{tall} matrices is smaller than on wide, while positive values indicate that error on \textit{wide} matrices is smaller than on \textit{tall}. The mean value (red line) denotes the mean of the distribution of differences.}
    \label{histograms_tall_wide}
\end{figure}

In our experiments, the $50$ generated synthetic $10 000 \times 300$ matrices are \textit{tall}, since the number of rows is greater than the number of columns. 
Moreover, we transpose each of the matrices to obtain a $300 \times 10 000$  \textit{wide} matrix. Since $(U \otimes V)^{\top}=V^{\top} \otimes U^{\top}$, transposing each matrix into a tall dataset is an equivalent but faster way of constructing a wide dataset compared to generating a new wide synthetic dataset.
For each method, we test tall and wide synthetic matrices to verify on which matrix shape a method returns a smaller approximation error.
The value $\omega$ reports the proportion of tests when a smaller error is achieved on the wide matrix compared to the tall matrix. The algorithm would not have a preferred matrix shape when $\omega=0.5$. Tall matrices perform better than wide matrices when $\omega=0$, while $\omega = 1$ indicates that the wide matrix shape is better than the tall matrix shape. In our case, the strategies \texttt{ByElement}, \texttt{ByMatrix} and \texttt{ByRow} achieve $\omega = 0.92$, $\omega = 0.92$ and $\omega = 1.0$, respectively.
We repeat the procedure for 100 synthetic datasets, 50 tall and 50 wide (for results on one dataset, see Figure ~\ref{NE_tall_vs_wide}).
For each method, we compute the distribution of the differences between the final errors on tall and wide matrices.
We used the mean value of the distribution (red vertical line) in Figure~\ref{histograms_tall_wide} to decide for each method whether it performs better on tall or wide matrices.
All the methods tested perform better on wide matrices. 
Note that \texttt{ByRow} strategy on wider matrices is closer to \texttt{ByMatrix} strategy, but on taller matrices closer to \texttt{ByElement} strategy. Hence we see the reason for \texttt{ByRow} strategy preferring wide matrices in choosing precision of the \texttt{ByMatrix} strategy over the fast but less accurate choice of indices in the \texttt{ByElement} strategy.
% intuition behind tall vs. wide
The another potential reason for the preference for wide over tall matrices in all three proposed strategies is the Random Acol initialization of matrix $U$. Note that Random Acol uses random columns of the data matrix $R$ to initialize matrix $U$, and it better transfers the variance from the data when selecting columns in the wide matrix than in the tall matrix, resulting in better initialization. In the case of a limited running time on large data matrices, the initialization step is crucial, making the wide matrix the preferred shape for all proposed strategies.

% graphs tall vs wide

% histogram

\subsubsection{Selecting the \texttt{FastSTMF} method on validation data}

Given the results on synthetic training data, we include a step for checking the matrix shape at the beginning. Each of the methods first inspects the shape of the matrix and, if necessary, transposes the matrix to a \textit{wider} shape.

\begin{figure}[H]
    \centering
    \includegraphics{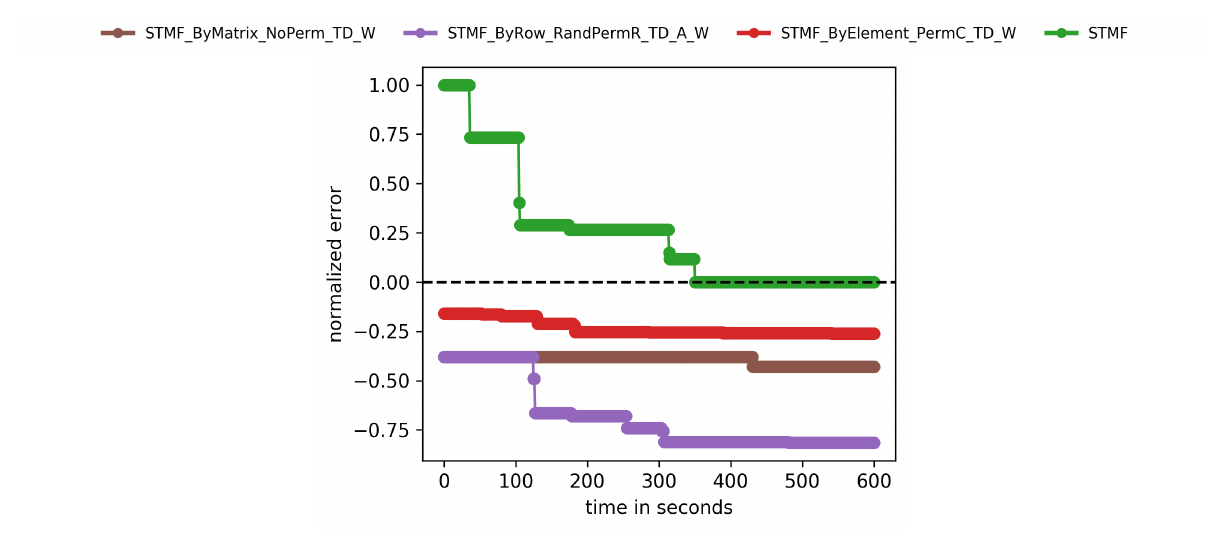}
    \caption{Median $\NE$ between methods on one large synthetic test rank three matrix of size $10 000 \times 300$. Note that \texttt{STMF} does not use matrix transposition, so it has a larger error at the beginning compared to other methods.}
    \label{graph_test_data}
\end{figure}

We rank the methods by the time needed to achieve \texttt{STMF}'s final $\NE_{t_{\max}}$, see Figure~\ref{graph_test_data}.
If the method never achieves the final \texttt{STMF}'s $\NE_{t_{\max}}$, it gets the lowest ranking, and if the two methods achieve error at the same time, they share the ranking position.

\begin{figure}[H]
    \centering
    \includegraphics{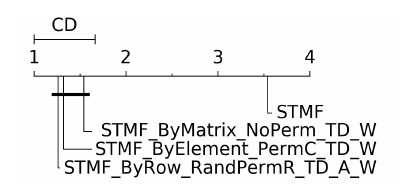}
    \caption{Critical difference (CD) graph of average performance ranks from Table~\ref{table_avg_rankings} for different methods tested on 50 large validation datasets, for $\alpha = 0.05$.}
    \label{critical_difference}
\end{figure}

When we compare the three proposed methods with \texttt{STMF} on synthetic validation data, we conclude that there are two groups of methods that differ significantly, see CD graph in Figure~\ref{critical_difference}.
The first group consists of our three proposed methods, and there is no significant difference between them. The second group contains only the original \texttt{STMF} method.
The average ranking on $50$ datasets is the highest for the \texttt{STMF\_ByRow\_RandPermR\_TD\_A\_W} method (see~Algorithm~\ref{pseudocode_fast_stmf}), which we propose as \texttt{FastSTMF} (Table~\ref{table_avg_rankings}). 

\begin{table}[H]
    \centering
    \scalebox{0.8}{
    \begin{tabular}{c|c}
        & \textbf{average ranking}\\
        \hline
        \texttt{STMF} & 3.54\\
        \texttt{STMF\_ByElement\_PermC\_TD\_W}  & 1.32\\
        \texttt{STMF\_ByRow\_RandPermR\_TD\_A\_W} (\texttt{FastSTMF}) & 1.26\\ 
        \texttt{STMF\_ByMatrix\_NoPerm\_TD\_W} & 1.54\\
        \hline
    \end{tabular}}
    \caption{The average ranking of time needed to achieve \texttt{STMF}'s final $\NE_{t_{\max}}$ for proposed methods on 50 large synthetic validation datasets.}
    \label{table_avg_rankings}
\end{table}

\subsection{Real data}
\label{sec:real_data}
% small matrices
In Table~\ref{table_merged_100}, we present the results on the predicted performance, measured with DC, RMSE-P, and RMSE-A, for small datasets. We have already shown~\cite{omanovic2021sparse} that on six out of nine datasets, \texttt{STMF} outperforms \texttt{NMF} according to DC. Most DC values are improved by \texttt{FastSTMF} and therefore \texttt{FastSTMF} outperforms \texttt{STMF} and \texttt{NMF} on most datasets. Moreover, \texttt{FastSTMF} also achieves a smaller RMSE-A compared to \texttt{STMF}, but \texttt{NMF} still achieves the smallest RMSE-A in most cases. However, in most cases, RMSE-P is the highest for \texttt{NMF}, indicating its tendency to overfit. \texttt{STMF} and \texttt{FastSTMF} do not have this issue, which we can see from the fact that the differences in RMSE-A and RMSE-P are minor.

\begin{table}[H]
    \centering
    \scalebox{0.7}{
    \begin{tabular}{c|c|c|c|c|c|c|c|c|c|c}
        \textbf{Metric} & \textbf{Method} & \textbf{AML} & \textbf{COLON} & \textbf{GBM} & \textbf{LIHC} & \textbf{LUSC} & \textbf{OV} & \textbf{SARC} & \textbf{SKCM} & \textbf{BIC} \\
        
        \hline
         %& \texttt{STMF\_W} & 0.53 & 0.55 & 0.78 & 0.57 & 0.60 & 0.64 & 0.43 & 0.46 & 0.45\\
          & \texttt{STMF} & \textbf{0.83*} & 0.65 & 0.69 & 0.49 & 0.53 & 0.56 & 0.59 & 0.55 & 0.35\\
        DC & \texttt{FastSTMF} & 0.74 & \textbf{0.69*} & \textbf{0.82*} & \textbf{0.60*} & \textbf{0.68} & \textbf{0.72*} & \textbf{0.62} & \textbf{0.59} & \textbf{0.46*} \\
         & \texttt{NMF} & 0.69 & 0.60 & 0.33 & 0.31 & 0.72* & 0.34 & 0.67* & 0.64* & 0.23\\
         
        \hline
         %& \texttt{STMF\_W} & 3.25 & 2.86 & 0.34 & 2.64 & 3.49 & 1.69 & 2.20 & 2.34 & 1.90\\
         & \texttt{STMF} & 2.55 & 2.30 & 0.32 & 2.63 & 2.62 & 1.71 & 2.19 & 2.43 & 2.07\\
        RMSE-P & \texttt{FastSTMF} & \textbf{2.49*} & \textbf{2.08*} & \textbf{0.22*} & \textbf{2.35*} & \textbf{2.16} & \textbf{1.46*} & \textbf{1.92*} & \textbf{2.02*} & \textbf{1.88*} \\
        & \texttt{NMF} & 2.60 & 2.54 & 0.76 & 2.58 & 2.10* & 2.73 & 2.47 & 2.45 & 2.91\\
        
        \hline
        %& \texttt{STMF\_W} & 3.09 & 2.78 & 0.34 & 2.68 & 3.43 & 1.66 & 2.17 & 2.41 & 1.84\\
        & \texttt{STMF} & 2.35 & 2.32 & 0.31 & 2.69 & 2.62 & 1.67 & 2.14 & 2.40 & 1.99\\
        RMSE-A & \texttt{FastSTMF} & \textbf{2.31} & \textbf{2.05} & \textbf{0.22*} & \textbf{2.39} & \textbf{2.11} & \textbf{1.42*} & \textbf{1.88} & \textbf{2.05} & \textbf{1.80}\\
        & \texttt{NMF} & 1.74* & 1.71* & 0.58 & 1.82* & 1.69* & 1.54 & 1.61* & 1.68* & 1.40*\\
        \hline
    \end{tabular}}
    \caption{Median DC, median RMSE-P and median RMSE-A of final (at $t_{\max}$) approximation matrix on gene expression data on small datasets. Result of the best method in the comparison between \texttt{STMF} and \texttt{FastSTMF} is shown in bold. Best result among all three methods (\texttt{STMF}, \texttt{FastSTMF}, \texttt{NMF}) are indicated by asterisk.}
\label{table_merged_100}
\end{table}

\begin{figure}[H]
    \centering
    \includegraphics{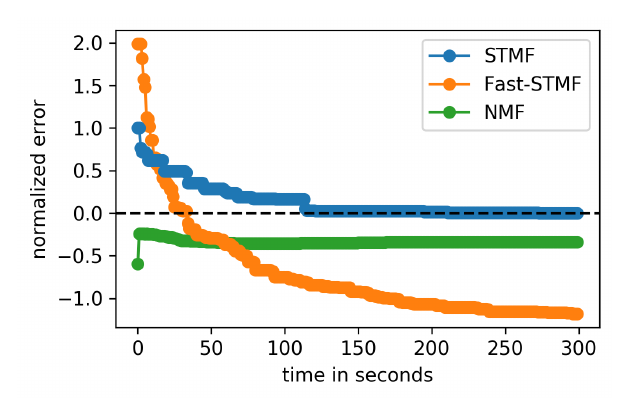}
    \caption{Comparison of median $\NE$ between \texttt{STMF}, \texttt{FastSTMF} and \texttt{NMF} on OV data.}
    \label{real_data_bic}
\end{figure}

Figure~\ref{real_data_bic} shows the results on the OV data where we can see that \texttt{FastSTMF} achieves smaller final $\NE_{t_{\max}}$ compared to \texttt{STMF} and \texttt{NMF}. Even more, \texttt{FastSTMF} reaches smaller $\NE$ than \texttt{STMF}'s in dozen of seconds and \texttt{NMF}'s final $\NE$ in sixty seconds. As we expected, \texttt{NMF} converges quickly, but it does not improve over time. For all other small datasets, similar figures are available in Supplement, Section~\ref{small_datasets_supp}.

% large matrices
Table~\ref{table_merged_1000} includes DC, RMSE-P, and RMSE-A for large datasets. We can see that according to DC, \texttt{FastSTMF} outperforms \texttt{STMF} on most datasets, however their RMSE results are similar.
In all cases except GBM, where the winner is \texttt{FastSTMF}, the best results among the three methods are achieved by \texttt{NMF}.
This behavior is very different from the one computed on small datasets. We believe that the reason behind this is that \texttt{NMF} tends towards the mean value of the data matrix entries, while \texttt{FastSTMF} and \texttt{STMF} can better fit to extreme values and distributions. Hence, the errors of the non-extremal entries are large, which in high-dimensional data matrices produces a larger RMSE of \texttt{FastSTMF} and \texttt{STMF} compared to \texttt{NMF}.

% merged table
\begin{table}[H]
    \centering
    \scalebox{0.7}{
    \begin{tabular}{c|c|c|c|c|c|c|c|c|c}
        \textbf{Metric} & \textbf{Method} & \textbf{AML} & \textbf{COLON} & \textbf{GBM} & \textbf{LIHC} & \textbf{LUSC} & \textbf{OV} & \textbf{SARC} & \textbf{SKCM} \\
        
        \hline
         & \texttt{STMF} & 0.53 & \textbf{0.54} & 0.66 & 0.50 & 0.49 & \textbf{0.49} & \textbf{0.48} & 0.50 \\
        DC & \texttt{FastSTMF} & \textbf{0.54} & 0.52 & \textbf{0.70*} & \textbf{0.53} & \textbf{0.56} & 0.46 & \textbf{0.48} & \textbf{0.59} \\
         & \texttt{NMF} & 0.86* & 0.87* & 0.66 & 0.83* & 0.86* & 0.84* & 0.91* & 0.84* \\ 
         
        \hline
         & \texttt{STMF} & 4.05 & \textbf{3.92} & 0.47 & \textbf{4.91} & 4.27 & \textbf{3.84} & 4.30 & \textbf{4.35} \\
        RMSE-P & \texttt{FastSTMF} & \textbf{3.97} & 3.99 & \textbf{0.46*} & 4.97 & \textbf{4.26} & 3.91 & \textbf{4.29} & 4.45 \\
        & \texttt{NMF} & 2.12* & 2.05* & 0.61 & 2.35* & 2.13* & 2.06* & 2.19* & 2.25* \\ 
        
        \hline
        & \texttt{STMF} & 4.04 & \textbf{3.92} & 0.47 & \textbf{4.80} & 4.26 & \textbf{3.85} & 4.32 & \textbf{4.33} \\
        RMSE-A & \texttt{FastSTMF} & \textbf{3.95} & 4.00 & \textbf{0.46*} & 4.95 & \textbf{4.25} & 3.91 & \textbf{4.30} & 4.43 \\
        & \texttt{NMF} & 2.01* & 1.97* & 0.56 & 2.30* & 2.07* & 1.97* & 2.12* & 2.18* \\ 
        \hline
        
    \end{tabular}}
    
    \caption{Median DC, median RMSE-P and median RMSE-A of final (at $t_{\max}$) approximation matrix on gene expression data on large datasets. Result of best method in the comparison between \texttt{STMF} and \texttt{FastSTMF} shown in bold. Best result among all three methods (\texttt{STMF}, \texttt{FastSTMF}, \texttt{NMF}) indicated by asterisk.}
    \label{table_merged_1000}
\end{table}

\begin{figure}[htb] % a and b part
    \centering
    \includegraphics{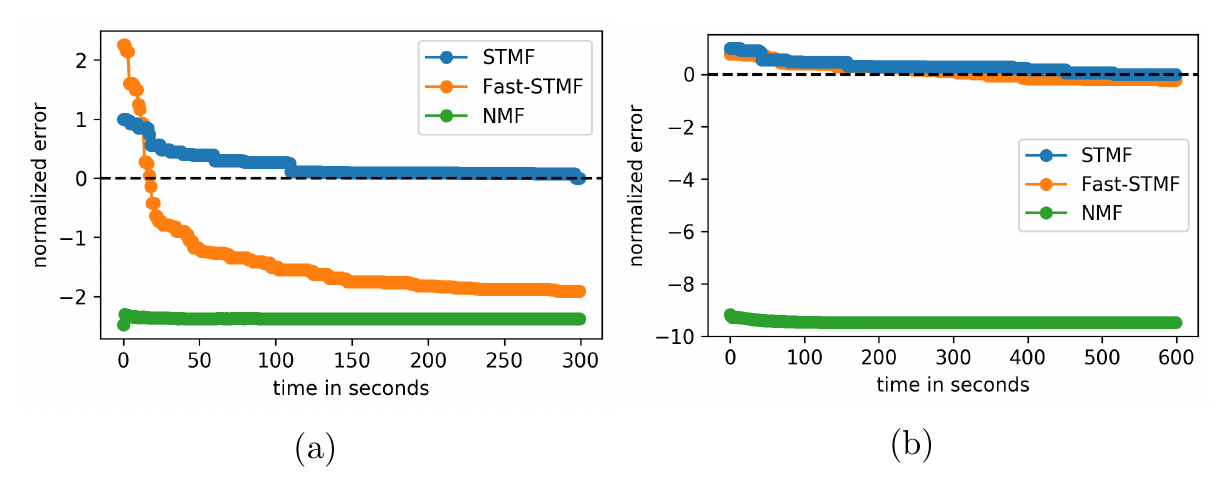}
    \caption{Comparison of median $\NE$ between \texttt{STMF}, \texttt{FastSTMF} and \texttt{NMF} on Sarcoma (SARC) data on: (a) small SARC data and (b) large SARC data.}
    \label{real_data_sarc}
\end{figure}

Figure~\ref{real_data_sarc} shows the results on the small and large SARC data where we can see that \texttt{FastSTMF} achieves a smaller final $\NE_{t_{\max}}$ faster than \texttt{STMF}.
Moreover, \texttt{FastSTMF} reaches smaller $\NE$ than \texttt{STMF}'s final $\NE_{t_{\max}}$ in dozen of seconds, see Figure~\ref{real_data_sarc}(a).
\texttt{NMF} converges quickly and achieves the smallest $\NE$ compared to other methods since it also achieves the smallest RMSE-A in Tables~\ref{table_merged_100} and~\ref{table_merged_1000}.
For all other large datasets, the figures are available in Supplement, Section~\ref{large_datasets_supp}.

\section{Conclusion}
Matrix factorization methods are one of the most widely used techniques in machine learning with various applications. The redefinition of basic operations, which mainly used standard linear algebra, enables to develop nonlinear models. However, all recent tropical semiring matrix factorization models have the main drawback in the slow optimization process, and in our work, we improve the speed of previous methods, while also achieving better results.

In this paper, we evaluate three strategies \texttt{ByElement}, \texttt{ByRow} and \texttt{ByMatrix} resulting in twelve methods tested on three types of structure in synthetic data: linear structure ($\lambda = 0$), tropical structure ($\lambda=1$) and their mixture ($\lambda = 0.5$). 
\texttt{ByRow} strategy achieves better results than \texttt{ByElement} and \texttt{ByMatrix} since its update is focused on only one entry in each row. \texttt{ByMatrix} strategy is extremely slow compared to other two strategies, since it firstly compares the errors of all matrix elements before the update. On the other extreme, \texttt{ByElement} iterates quickly, but it often chooses wrong elements to update since it chooses locally and does not consider the impact of the update on the approximation of the rest of the matrix.
\texttt{ByRow} strategy solves these problems by partitioning the matrix into rows and finding the best column and factor candidates that have the largest impact not only on the row, but also on the corresponding column.
In this way, the \texttt{ByRow} strategy achieves the best results by being faster than the \texttt{ByMatrix} strategy and more accurate than the \texttt{ByElement} strategy. For them we tested different $\td$-based heuristics that have different efficiencies for different choices of $\lambda$, so for a specific data pattern one might compare and select the most appropriate heuristic. We proved that \texttt{TD\_A} selection, which is more global than \texttt{TD} selection, gives the best results when there is no prior knowledge about $\lambda$.

We propose an algorithm based on tropical semiring called \texttt{FastSTMF}, which has proven to be an efficient matrix factorization method for the prediction of missing values (matrix completion task). \texttt{FastSTMF} is based on \texttt{STMF}, but it introduces a new way of updating factor matrices, which results in higher performance, \textit{i.e.}, achieving better results in a shorter time.
The results on synthetic data show that \texttt{FastSTMF} outperforms \texttt{STMF} in achieving a smaller approximation error in less time, while on real small data \texttt{FastSTMF} outperforms \texttt{STMF} and \texttt{NMF} in achieving a higher distance correlation and a smaller prediction and approximation error. The results on real large data confirm the superior performance of \texttt{FastSTMF} over \texttt{STMF} in terms of distance correlation, but it is shown that \texttt{NMF} outperforms both methods if we have high-dimensional data. 

A limitation of our \texttt{FastSTMF} method is its inability to achieve comparable results on high-dimensional data to methods that use standard linear algebra. 
In our future work, we should explore how to reduce the dimensionality of the data to select more suitable features for tropical optimization. Another drawback in the algorithm's speed is in the update procedure compared to methods that use standard linear algebra. \texttt{NMF} uses \textit{gradient descent} and updates \textit{all} factor matrices' values simultaneously. Since the gradient descent cannot be used in tropical semiring, we change the entries of only one column of the coefficient factor and only one row of the basis factor in one update step (\texttt{F-ULF} or \texttt{F-URF}).
With this, we guarantee the convergence of \texttt{FastSTMF}, but addressing the simultaneous update of more rows and columns of factor matrices in tropical semiring deserves further detailed research. 
Note that RMSE favors the normally distributed data and models that tend toward the mean approximation, such as \texttt{NMF}. On the other hand, \texttt{FastSTMF} prefers extreme values and hence RMSE penalizes the large errors of the non-extremal entries. We should explore which metrics could better compare different methods independently of preferred data distribution. 
Our ideas articulate a wide range of research questions of dependence of proposed methods on different probability distributions for generating synthetic datasets (\textit{e.g.}, beta, Poisson), and also the question whether substituting the $b$-norm in our objective function with the $\ell_\infty$-norm~\cite{maragos2021tropical}  %(max norm) 
would give better results.

This paper presents a novel and faster way to guide the optimization process in matrix factorization methods based on tropical semiring. We show how to identify the best method based on the patterns that appear in the data. We believe that the proposed \texttt{FastSTMF} and its variants are essential to develop other techniques based on general semirings, which could help to find different structures not explainable by standard linear algebra.

\section*{Author's contributions}
AO, TC and PO designed the study. AO wrote the software application and performed experiments. AO and TC analyzed and interpreted the results on real data. AO wrote the initial draft of the paper and all authors edited and approved the final manuscript.

\section*{Declaration of Competing Interest}
The authors declare that they have no known competing financial interests or personal relationships that could have appeared to influence the work reported in this paper.

\section*{Funding}
This work is supported by the Slovene Research Agency, Young Researcher Grant (52096) awarded to AO, and research core funding (P1-0222 to PO and P2-0209 to TC).

\section*{Availability of data and materials}
This paper uses the real TCGA data available on \url{http://acgt.cs.tau.ac.il/multi_omic_benchmark/download.html}. \texttt{PAM50} data can be found on the \url{https://github.com/CSB-IG/pa3bc/tree/master/bioclassifier_R/}. \texttt{BIC} subtypes are collected from \url{https://www.cbioportal.org/}. \texttt{STMF} code, \texttt{PAM50} data and \texttt{BIC} subtypes are available on \url{https://github.com/Ejmric/FastSTMF}.

\bibliographystyle{unsrt}  

%\bibliographystyle{plain}
\bibliography{bibliography}

\section*{Supplementary material}
\appendix

\renewcommand\figurename{Supplementary Figure S}

\section{Performance comparison}

\begin{figure}[H]
    \centering
    \captionsetup{justification=centering}
    \includegraphics[scale=0.3]{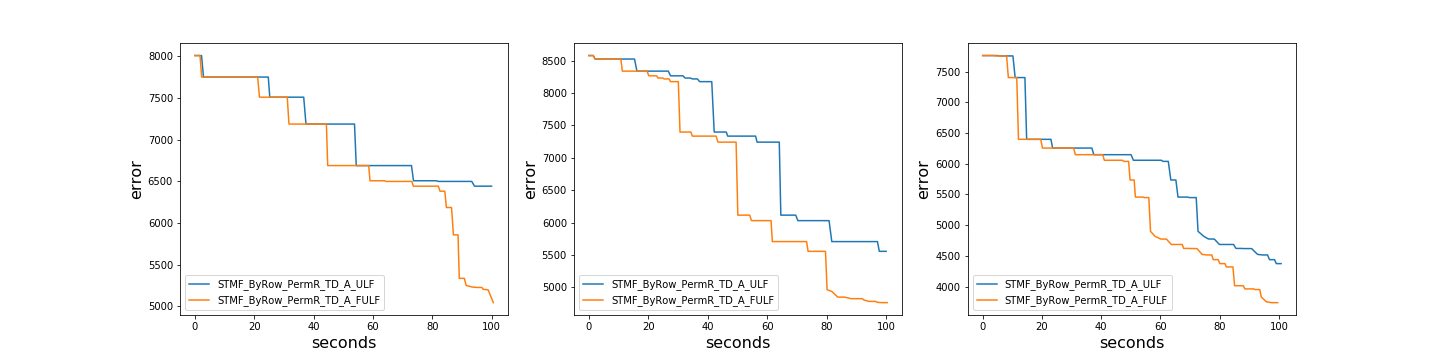}
    \caption{Difference between \texttt{ULF} and \texttt{F-ULF}.}
    \label{ulf_vs_fulf}
\end{figure}

\begin{figure}[H]
    \centering
    \captionsetup{justification=centering}
    \includegraphics[scale=0.3]{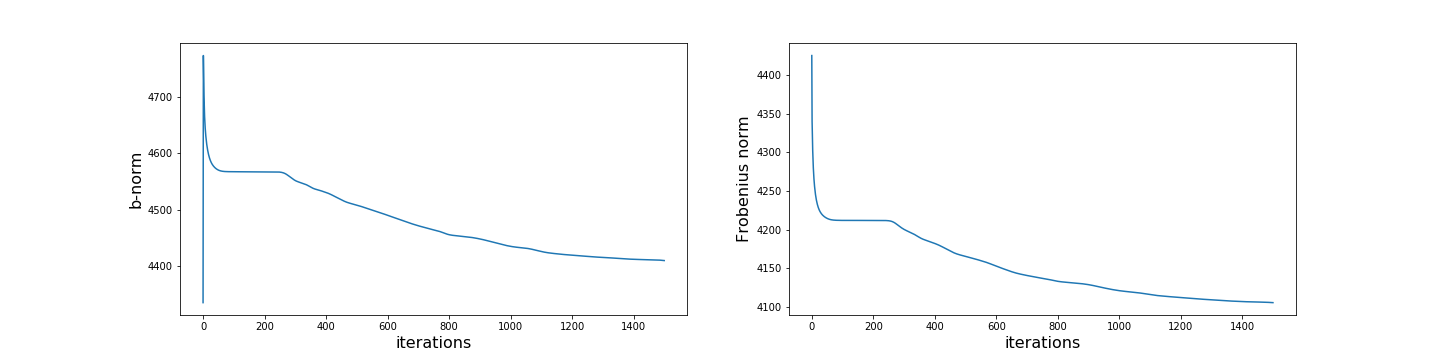}
    \caption{Comparison between Frobenius and $b$-norm norm in \texttt{NMF}.}
    \label{nmf_bnorm_vs_fro}
\end{figure}

\begin{figure}[htp]
     \centering
     \begin{subfigure}[t]{0.45\textwidth}
         \centering
         \includegraphics[width=\textwidth]{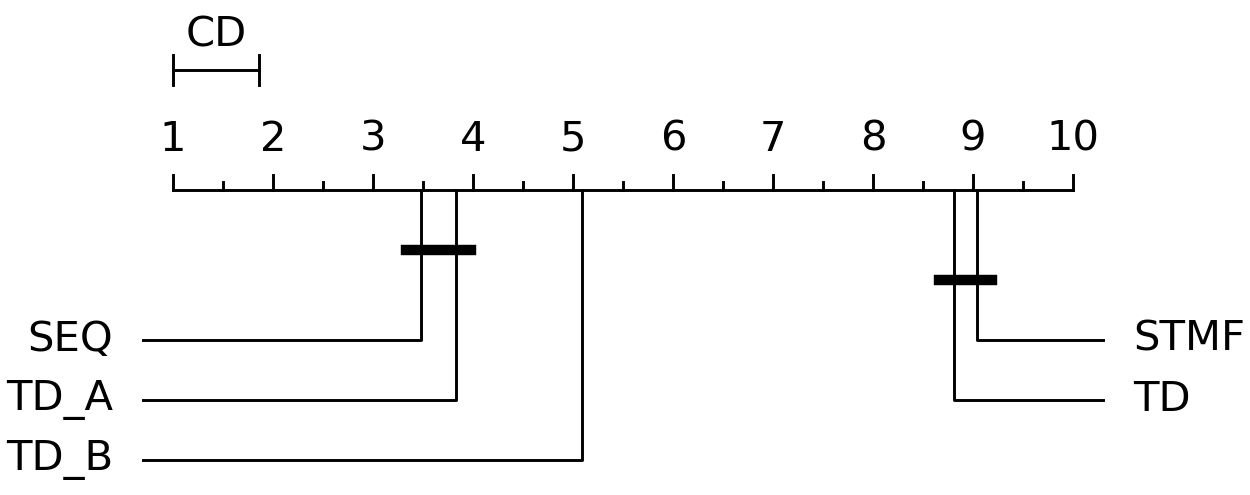}
         \caption{Ranking on time needed to achieve \texttt{STMF}'s final $\NE_{t_{\max}}$.}
     \end{subfigure}
     \hfill
     \begin{subfigure}[t]{0.45\textwidth}
         \centering
         \includegraphics[width=\textwidth]{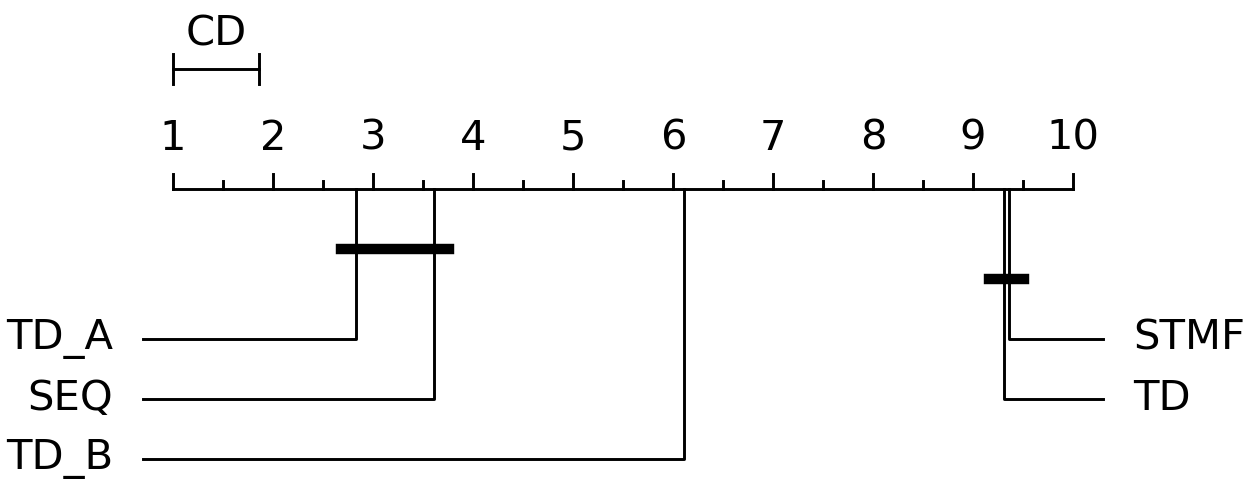}
         \caption{Ranking on final $\NE_{t_{\max}}$.}
     \end{subfigure}
        \caption{CD graphs for $\lambda = 0$. Average rankings of methods and critical difference (CD) at 0.05 significance level calculated on 50 datasets.}
        \label{supp_CD_lambda_0}
\end{figure}

\begin{figure}[htp]
     \centering
     \begin{subfigure}[t]{0.45\textwidth}
         \centering
         \includegraphics[width=\textwidth]{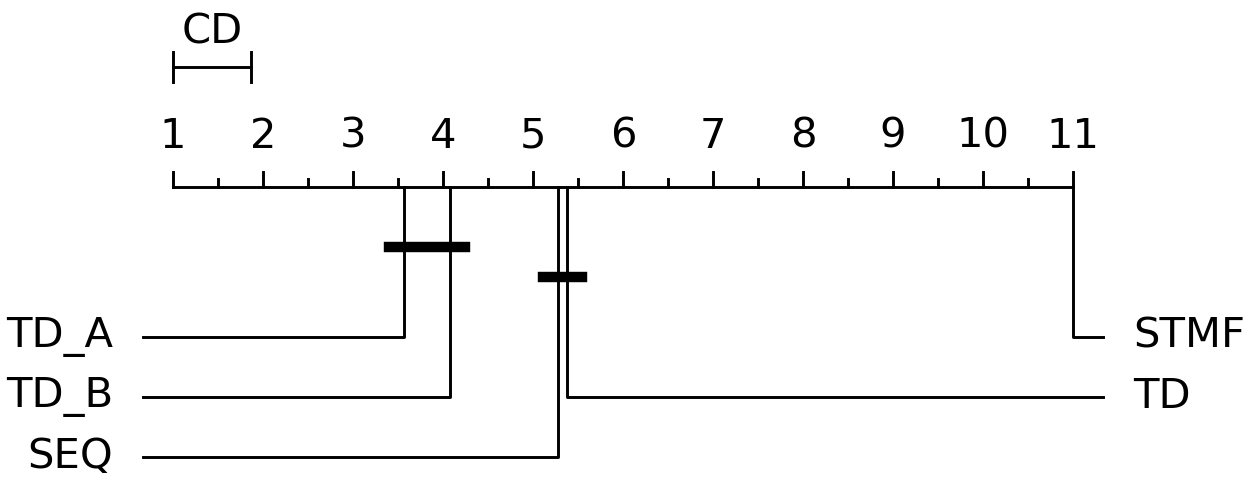}
         \caption{Ranking on time needed to achieve \texttt{STMF}'s final $\NE_{t_{\max}}$.}
     \end{subfigure}
     \hfill
     \begin{subfigure}[t]{0.45\textwidth}
         \centering
         \includegraphics[width=\textwidth]{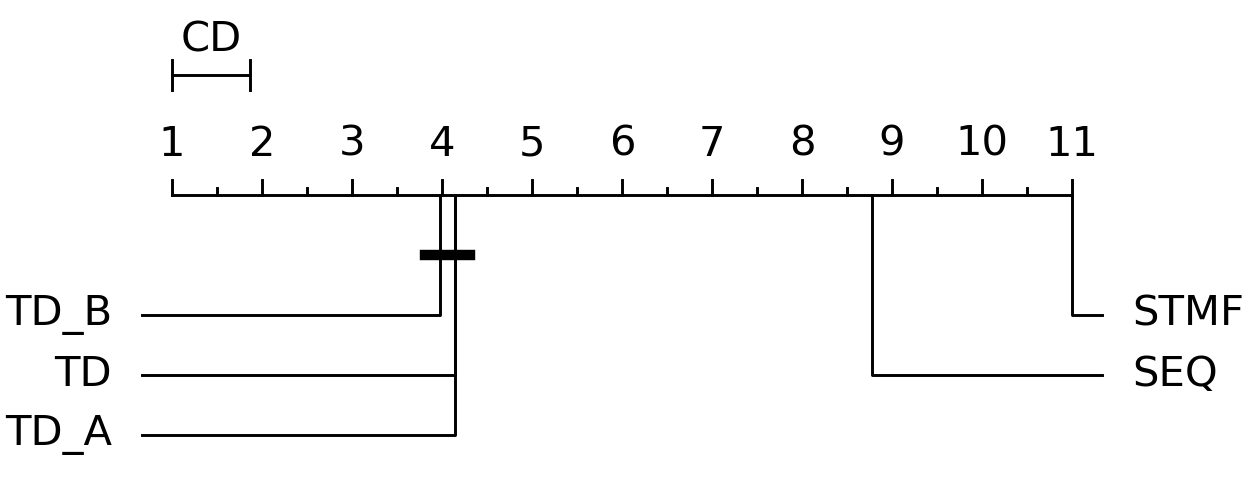}
         \caption{Ranking on final $\NE_{t_{\max}}$.}
     \end{subfigure}
        \caption{CD graphs for $\lambda = 0.5$. Average rankings of methods and critical difference (CD) at 0.05 significance level calculated on 50 datasets.}
        \label{supp_CD_lambda_0.5}
\end{figure}

\begin{figure}[htp]
     \centering
     \begin{subfigure}[t]{0.45\textwidth}
         \centering
         \includegraphics[width=\textwidth]{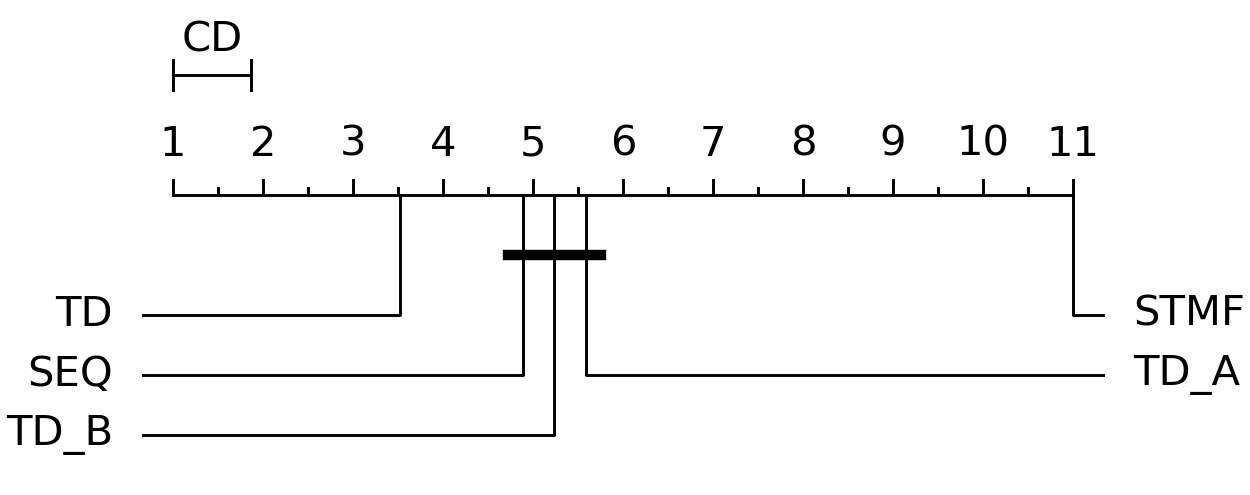}
         \caption{Ranking on time needed to achieve \texttt{STMF}'s final $\NE_{t_{\max}}$.}
     \end{subfigure}
     \hfill
     \begin{subfigure}[t]{0.45\textwidth}
         \centering
         \includegraphics[width=\textwidth]{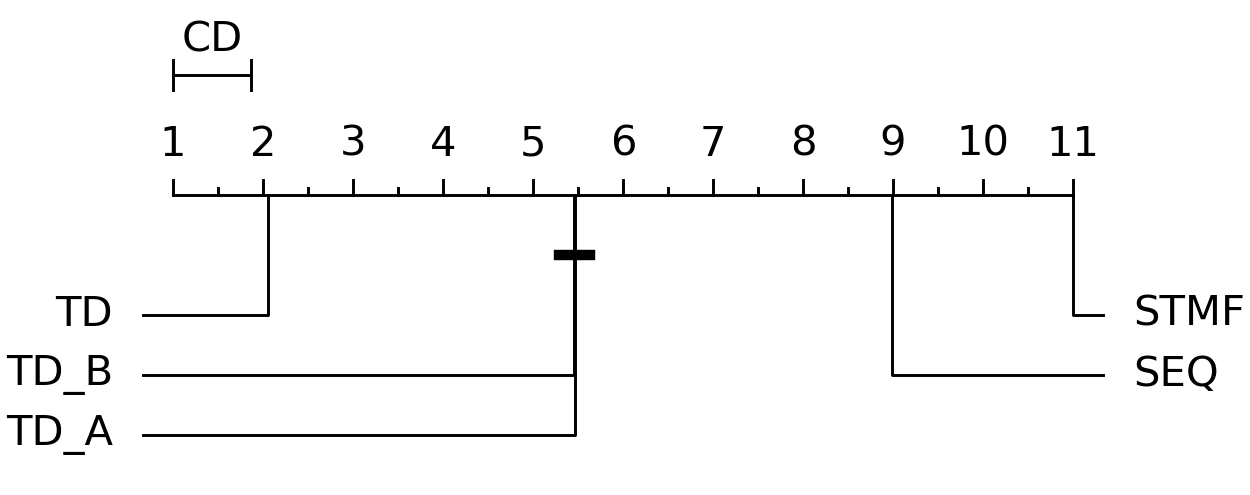}
         \caption{Ranking on final $\NE_{t_{\max}}$.}
     \end{subfigure}
        \caption{CD graphs for $\lambda = 1$. Average rankings of methods and critical difference (CD) at 0.05 significance level calculated on 50 datasets.}
        \label{supp_CD_lambda_1}
\end{figure}

\clearpage
\subsection{Small datasets, normalized errors of \texttt{Fast-STMF}, \texttt{STMF} and \texttt{NMF}}
\label{small_datasets_supp}

\begin{paracol}{2}
\begin{figure}[H]
    \centering
    \includegraphics[width=0.35\textwidth]{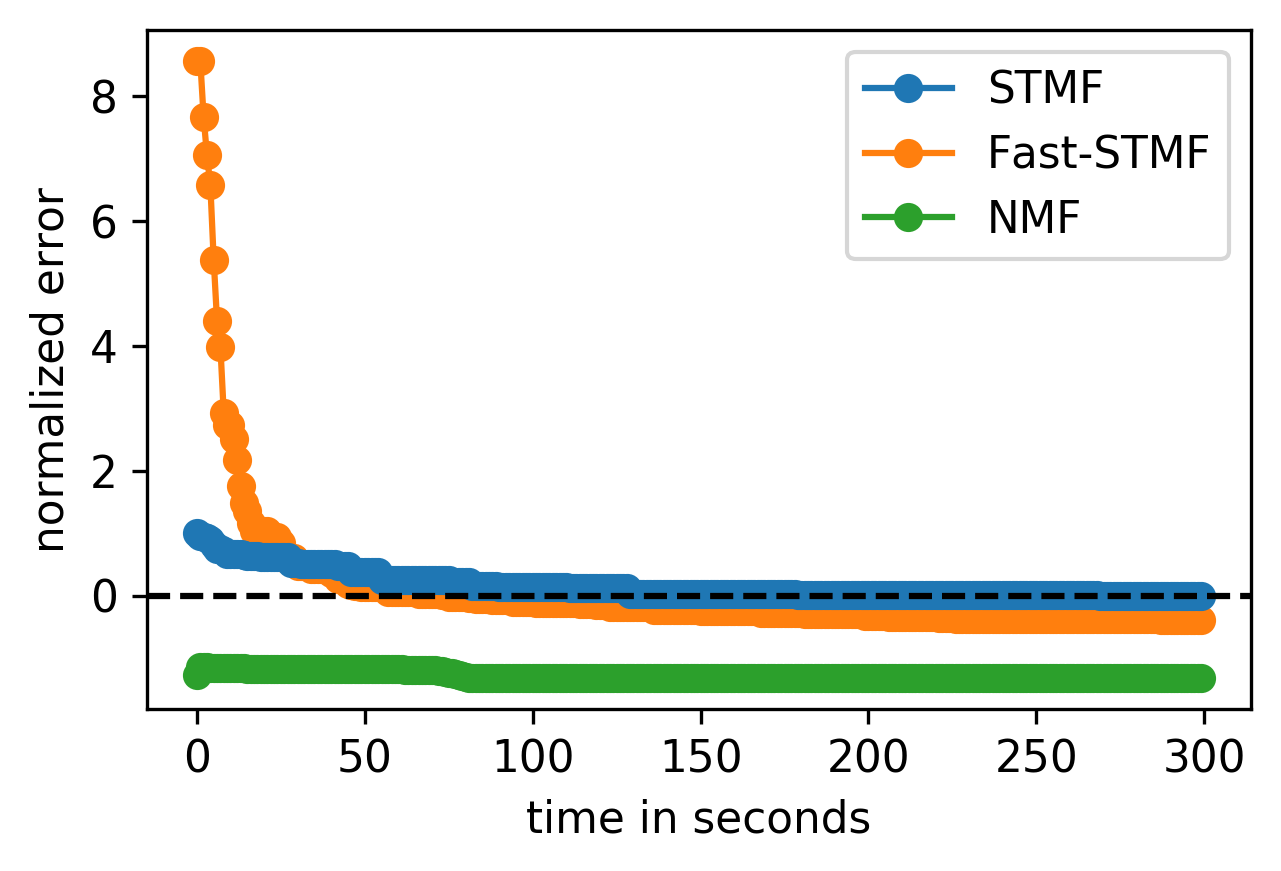}
    \caption{AML small}
    \label{small_aml}
\end{figure}
\switchcolumn
\begin{figure}[H]
    \centering      
    \includegraphics[width=0.35\textwidth]{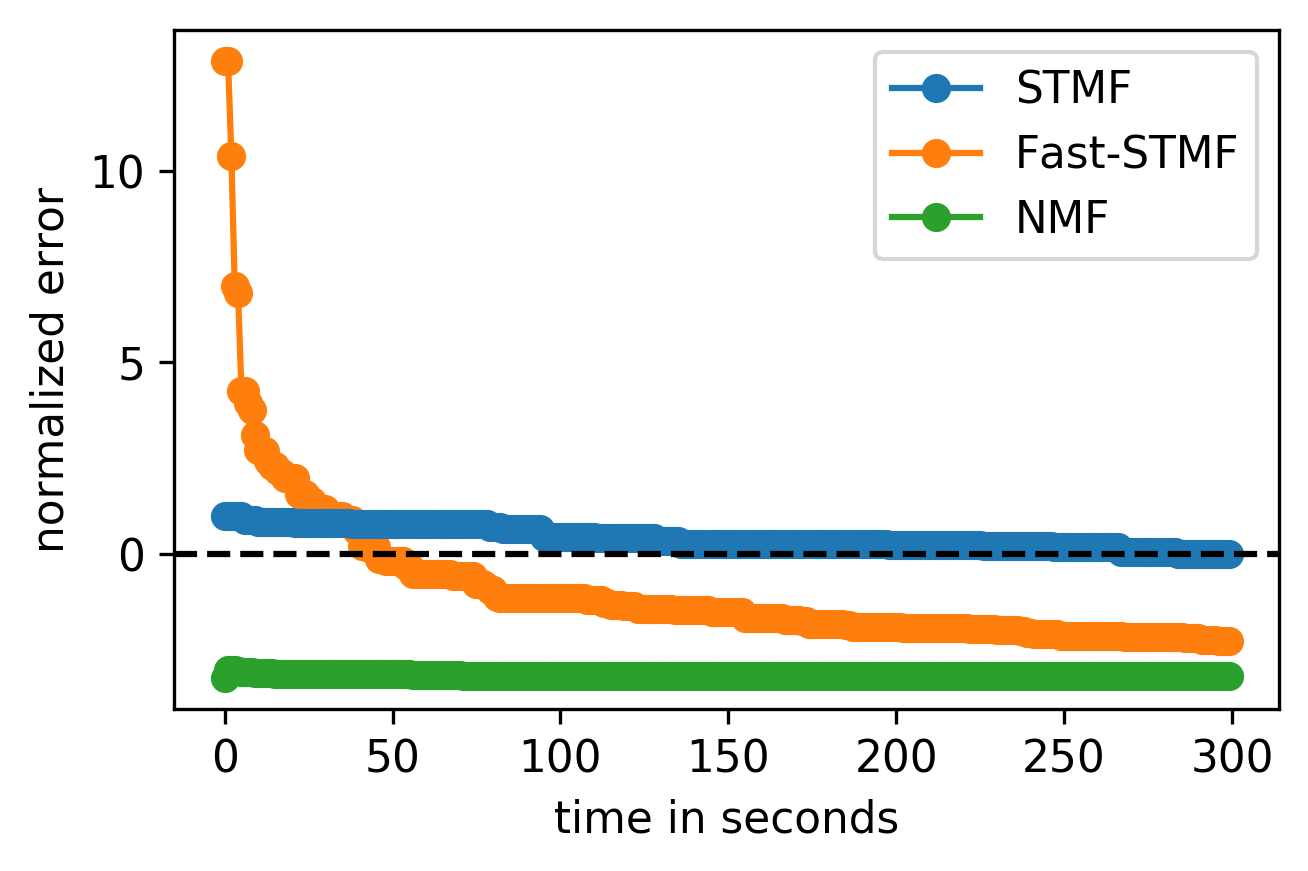}
    \caption{COLON small}
    \label{small_colon}
\end{figure}
\end{paracol}

\begin{paracol}{2}
\begin{figure}[H]
    \centering
    \includegraphics[width=0.35\textwidth]{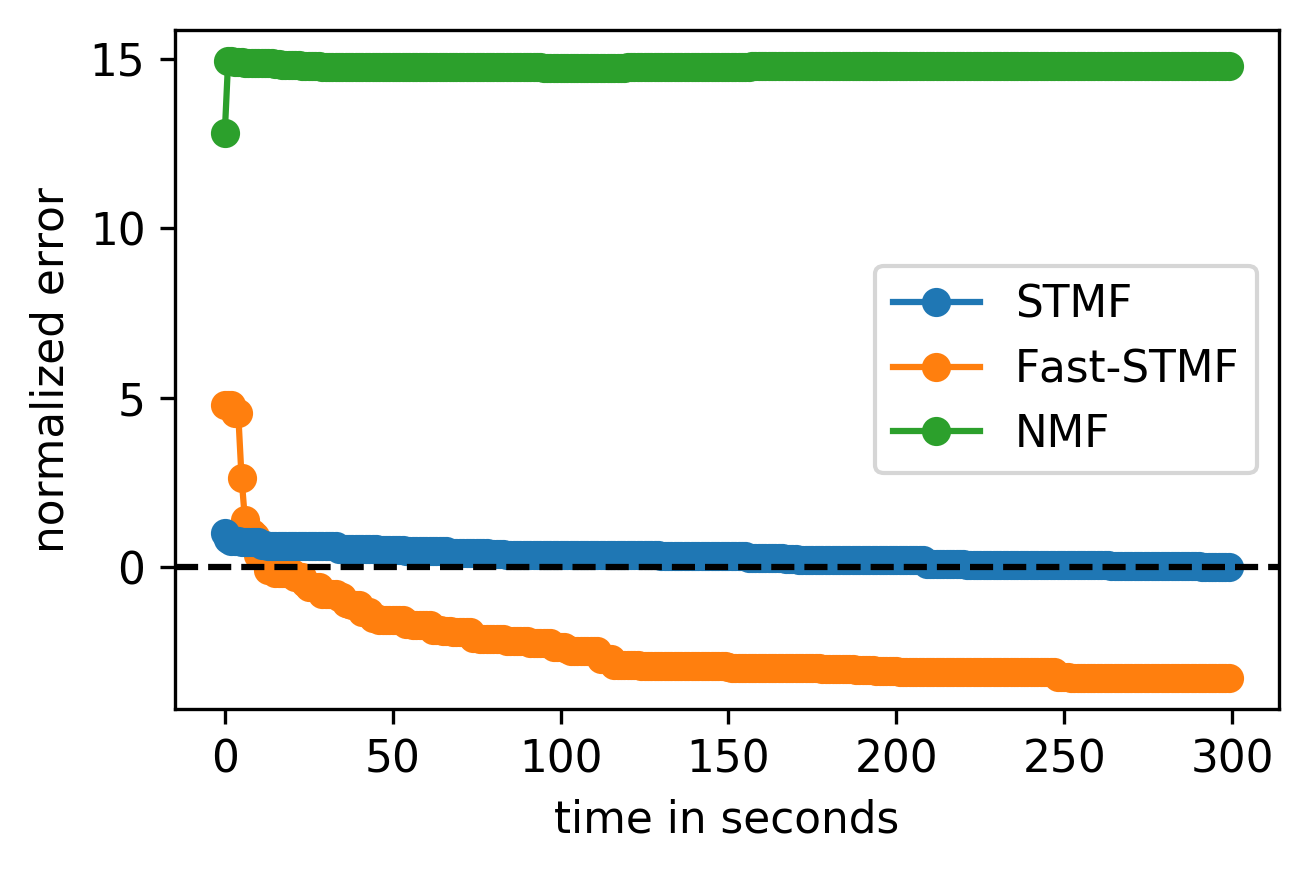}
    \caption{GBM small}
    \label{small_gbm}
\end{figure}
\switchcolumn
\begin{figure}[H]
    \centering      
    \includegraphics[width=0.35\textwidth]{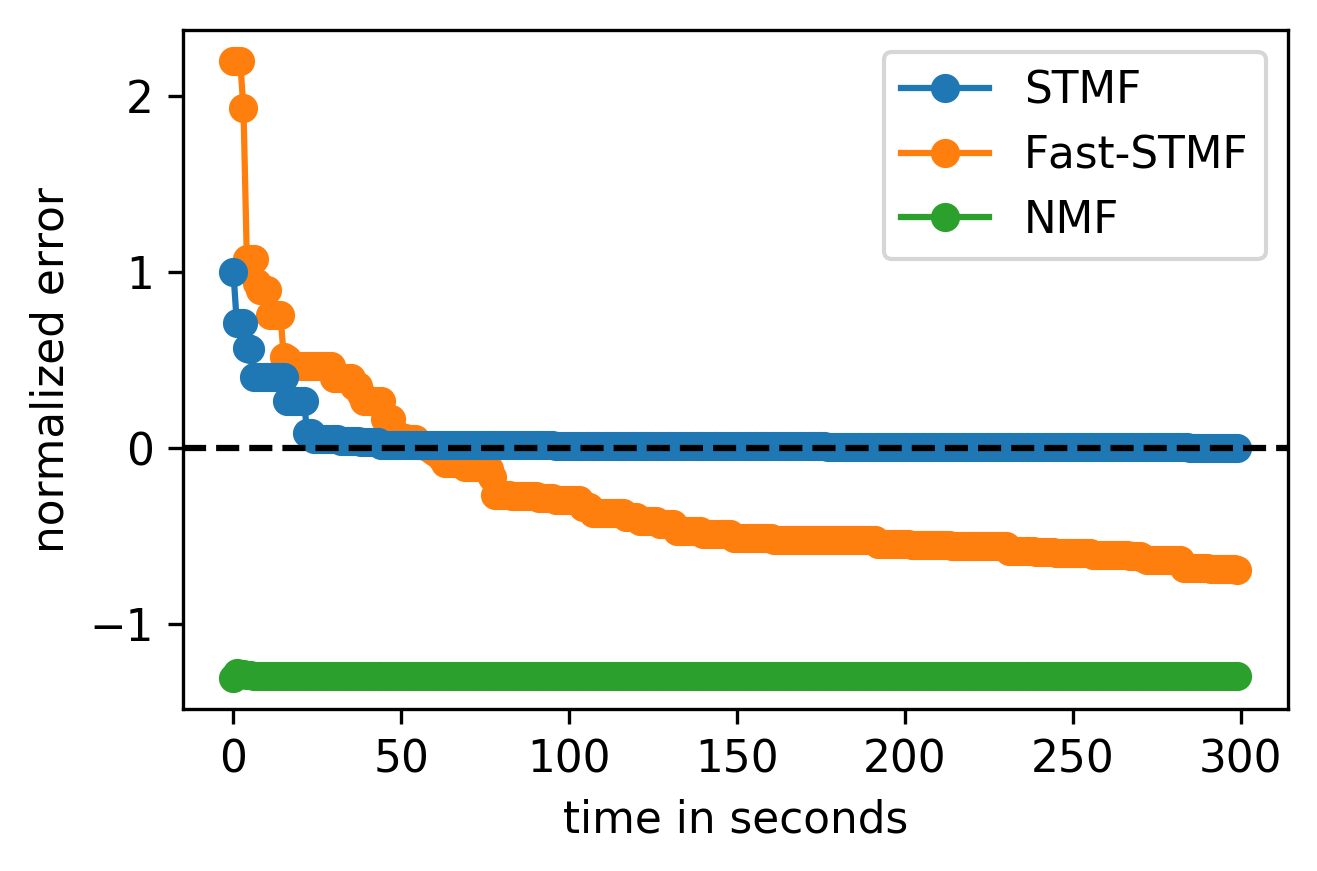}
    \caption{LIHC small}
    \label{small_lihc}
\end{figure}
\end{paracol}

\begin{paracol}{2}
\begin{figure}[H]
    \centering
    \includegraphics[width=0.35\textwidth]{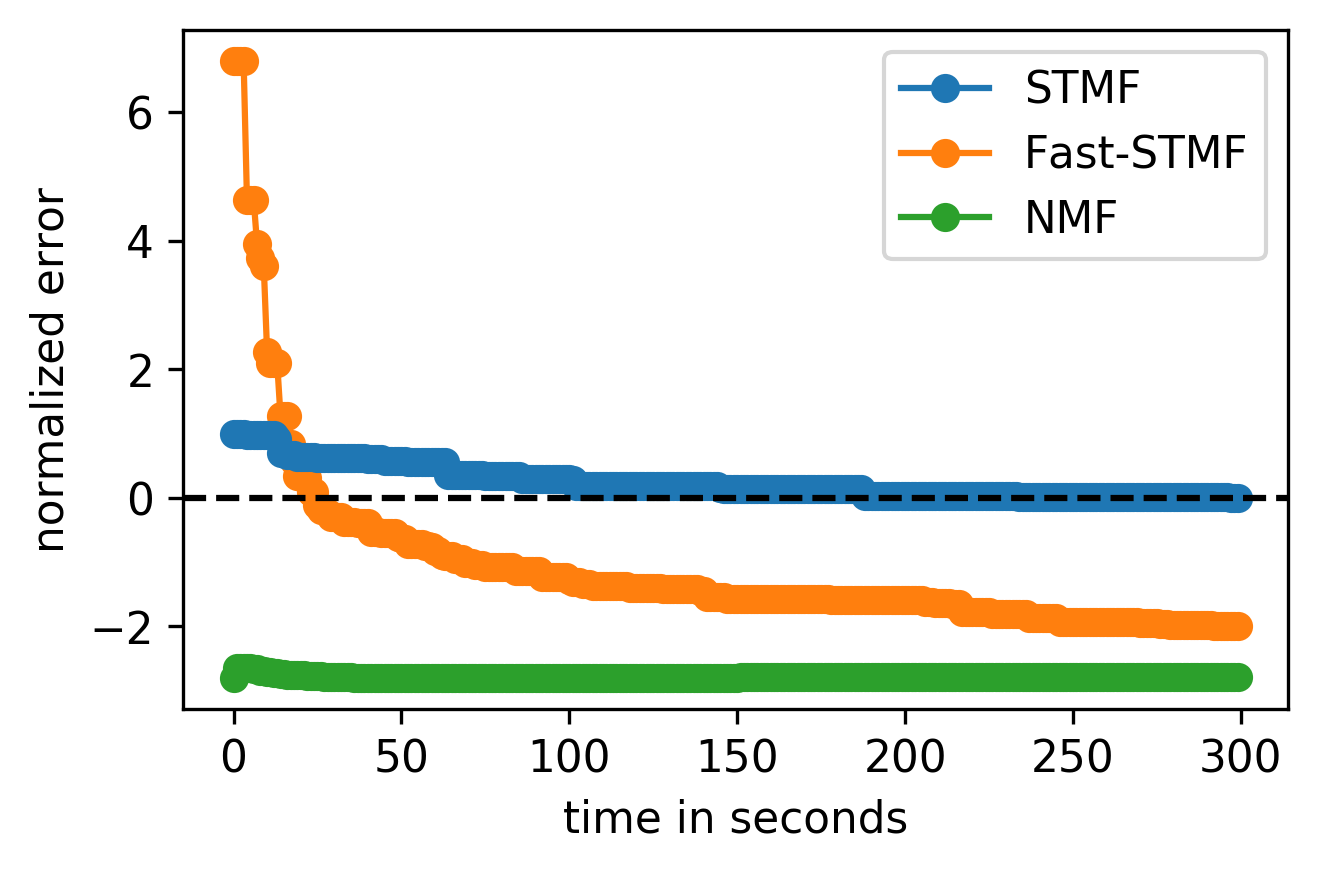}
    \caption{LUSC small}
    \label{small_lusc}
\end{figure}
\switchcolumn
\begin{figure}[H]
    \centering      
    \includegraphics[width=0.35\textwidth]{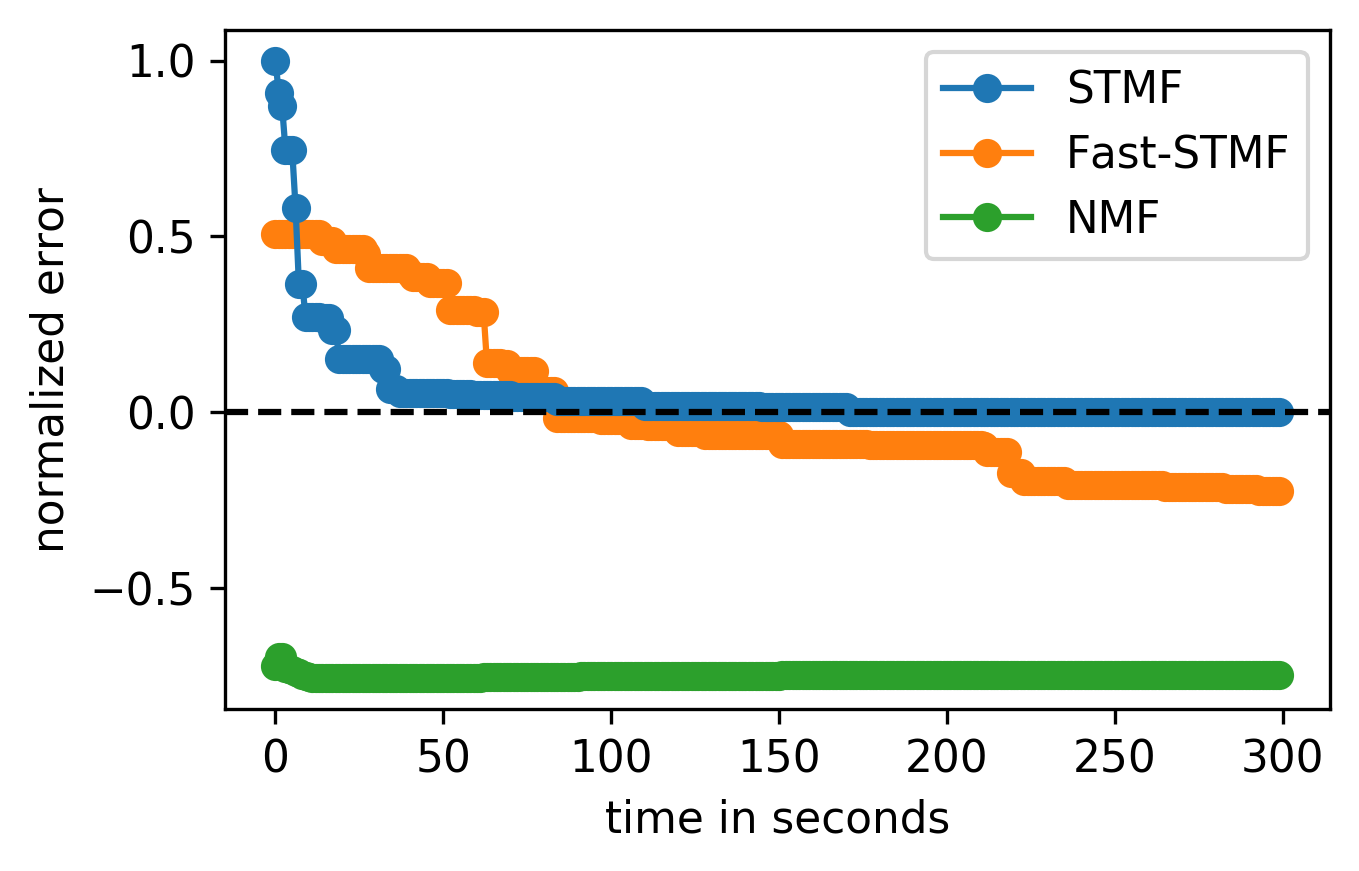}
    \caption{BIC small}
    \label{small_bic}
\end{figure}
\end{paracol}

\begin{paracol}{2}
\begin{figure}[H]
    \centering
    \includegraphics[width=0.35\textwidth]{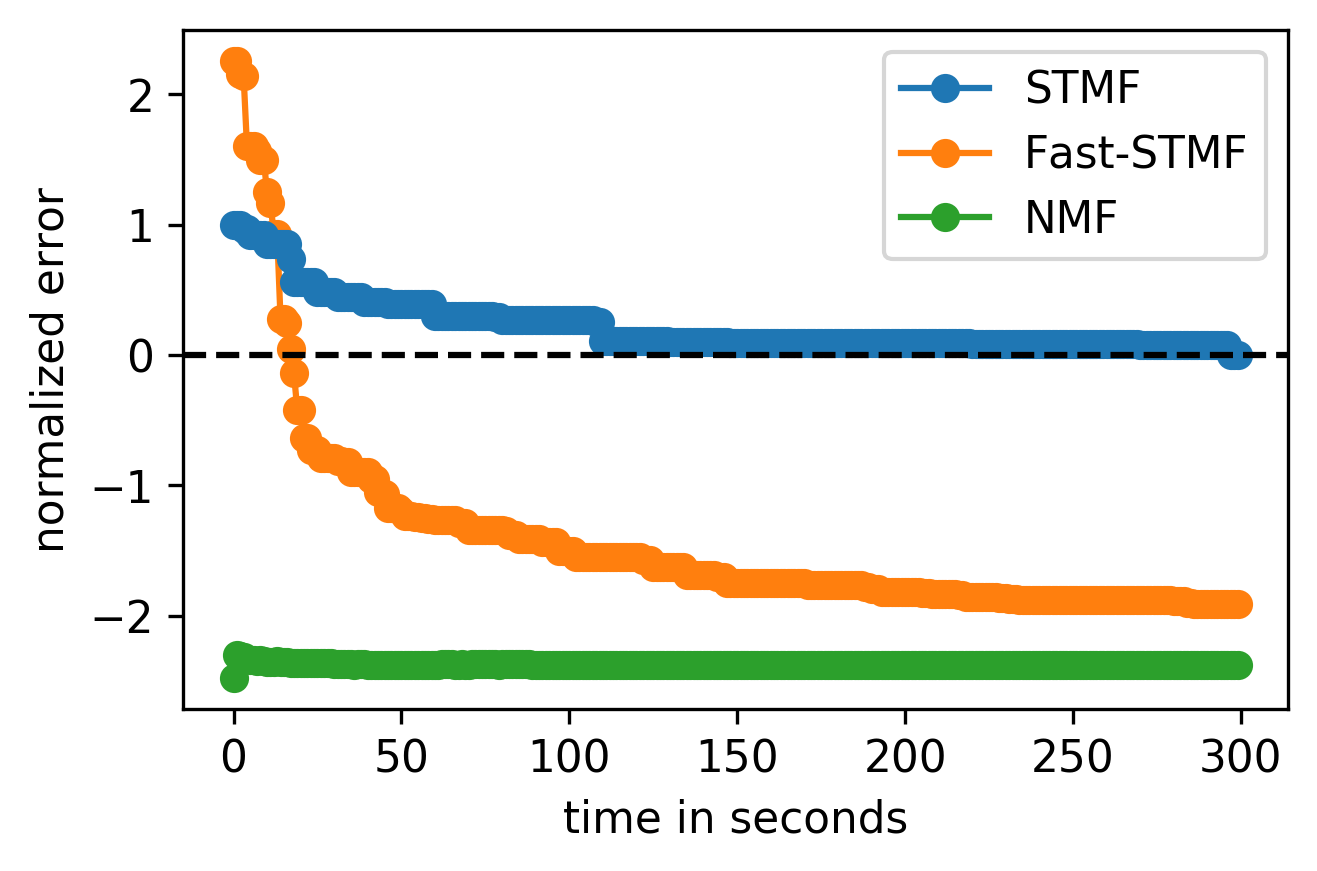}
    \caption{SARC small}
    \label{small_sarc}
\end{figure}
\switchcolumn
\begin{figure}[H]
    \centering      
    \includegraphics[width=0.35\textwidth]{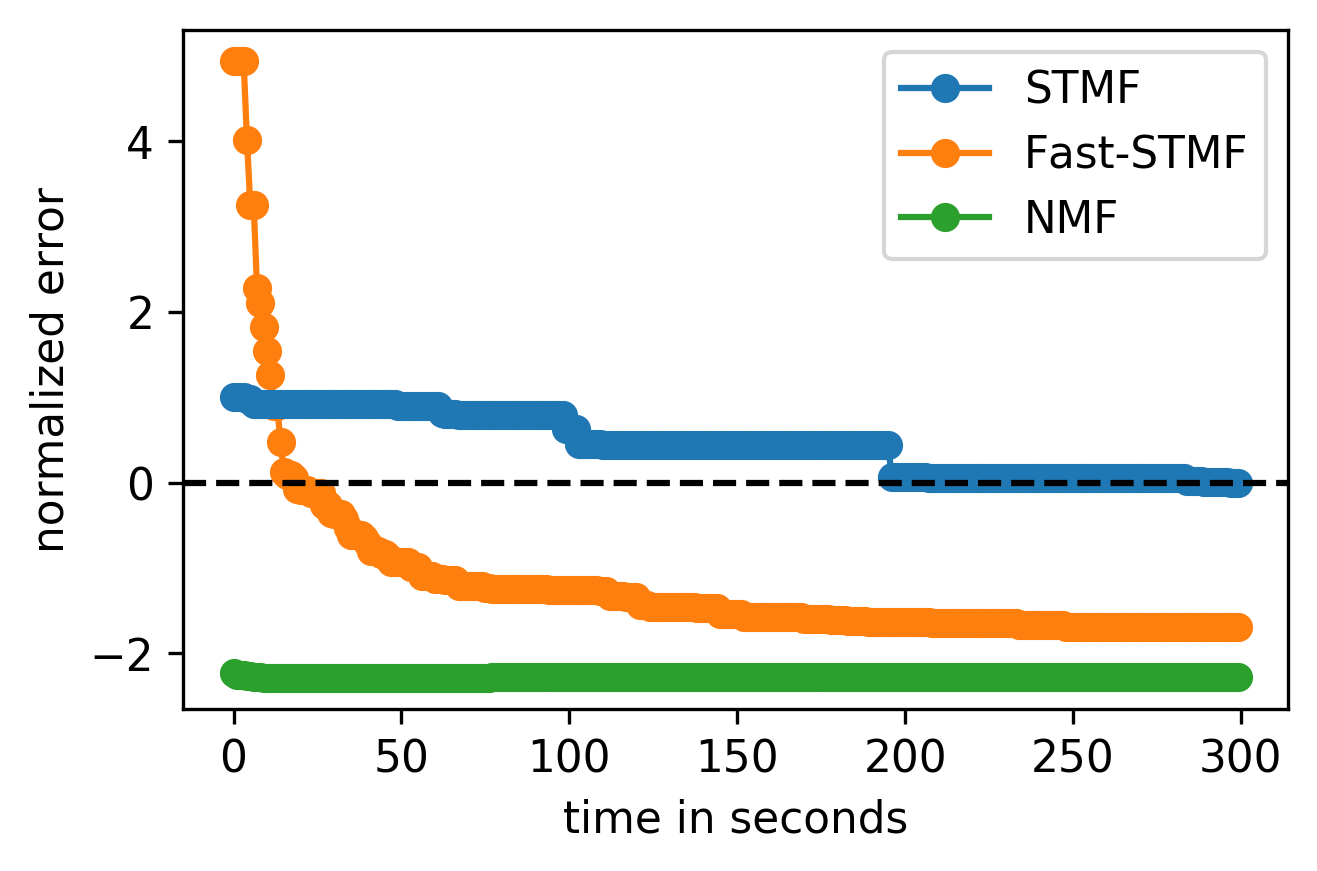}
    \caption{SKCM small}
    \label{small_skcm}
\end{figure}
\end{paracol}

\clearpage
%%%%%%%%%%%%%%%%%%%%%%%%%%%
\subsection{Large datasets, normalized errors of \texttt{Fast-STMF}, \texttt{STMF} and \texttt{NMF}}
\label{large_datasets_supp}

\begin{paracol}{2}
\begin{figure}[H]
    \centering
    \includegraphics[width=0.35\textwidth]{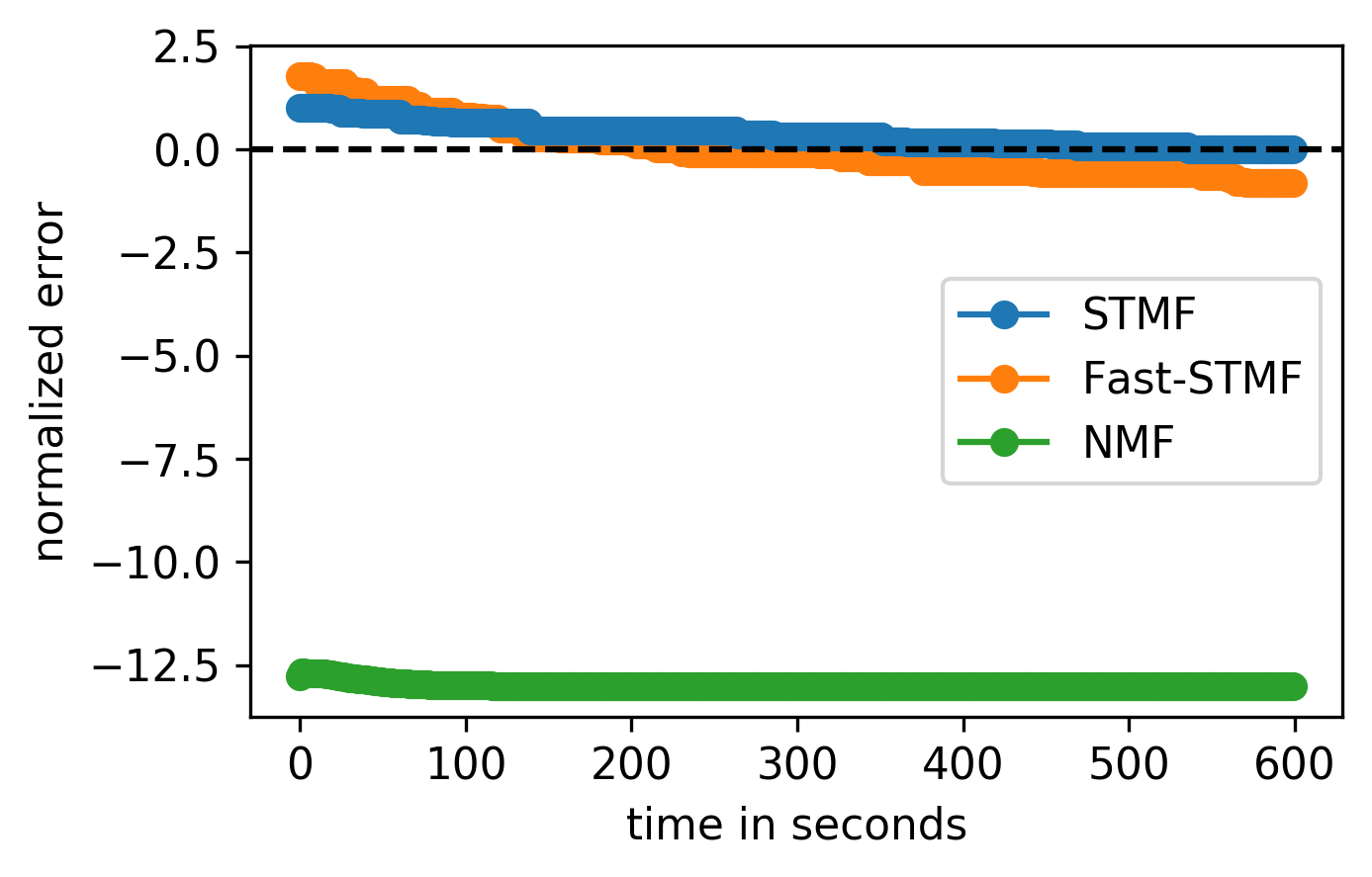}
    \caption{AML large}
    \label{large_aml_supp}
\end{figure}
\switchcolumn
\begin{figure}[H]
    \centering      
    \includegraphics[width=0.35\textwidth]{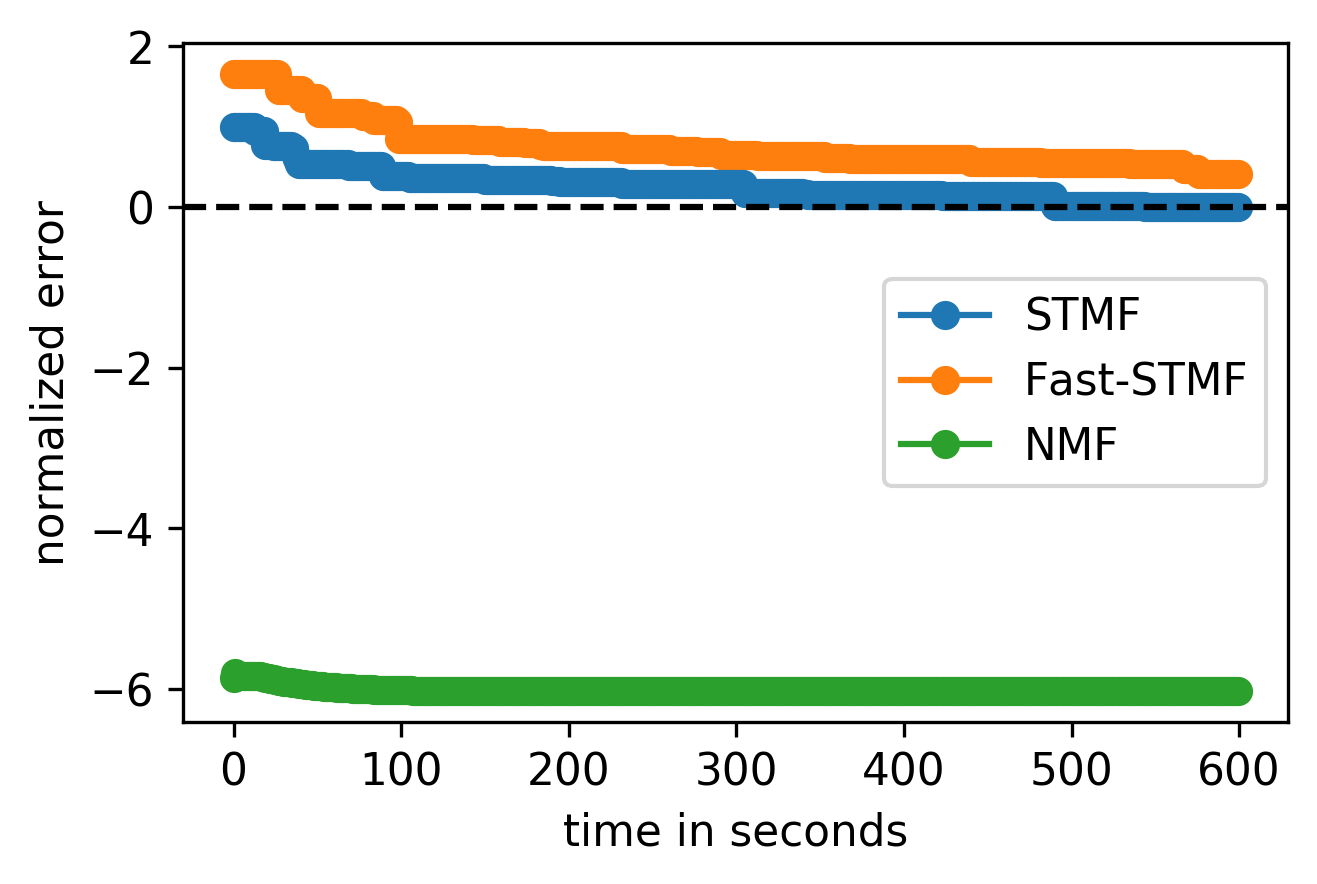}
    \caption{COLON large}
    \label{large_colon}
\end{figure}
\end{paracol}

\begin{paracol}{2}
\begin{figure}[H]
    \centering
    \includegraphics[width=0.35\textwidth]{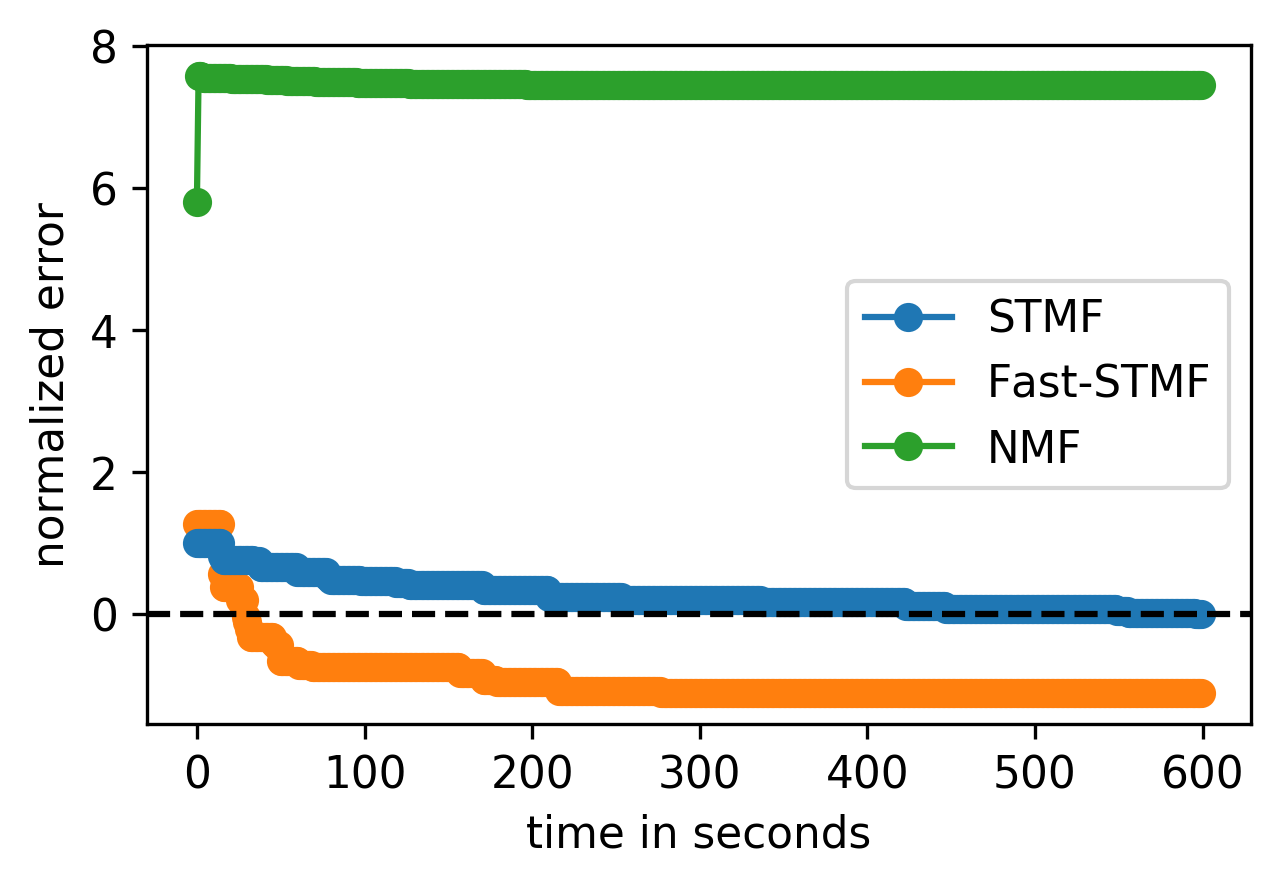}
    \caption{GBM large}
    \label{large_gbm}
\end{figure}
\switchcolumn
\begin{figure}[H]
    \centering      
    \includegraphics[width=0.35\textwidth]{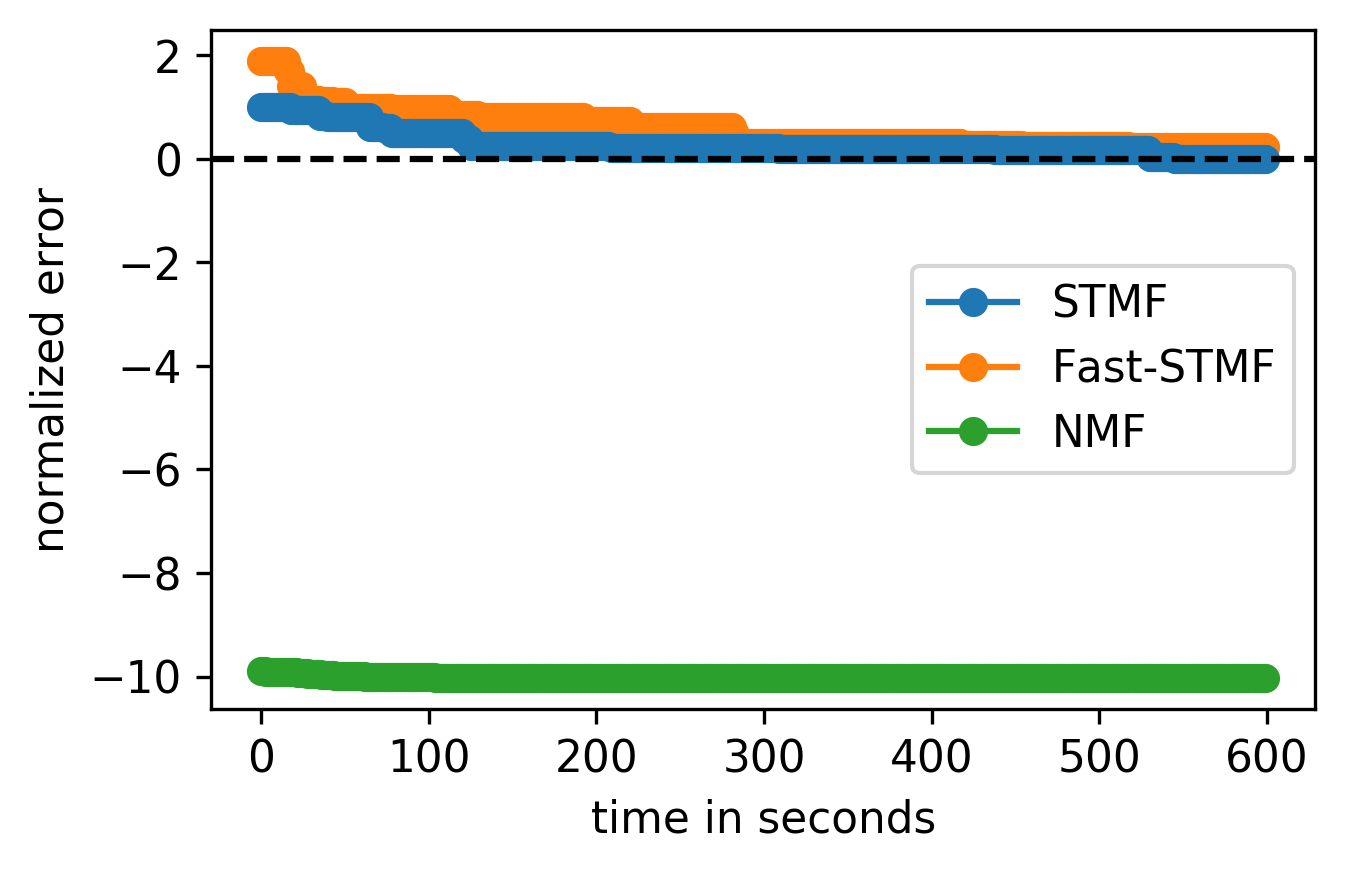}
    \caption{LIHC large}
    \label{large_lihc}
\end{figure}
\end{paracol}

\begin{paracol}{2}
\begin{figure}[H]
    \centering
    \includegraphics[width=0.35\textwidth]{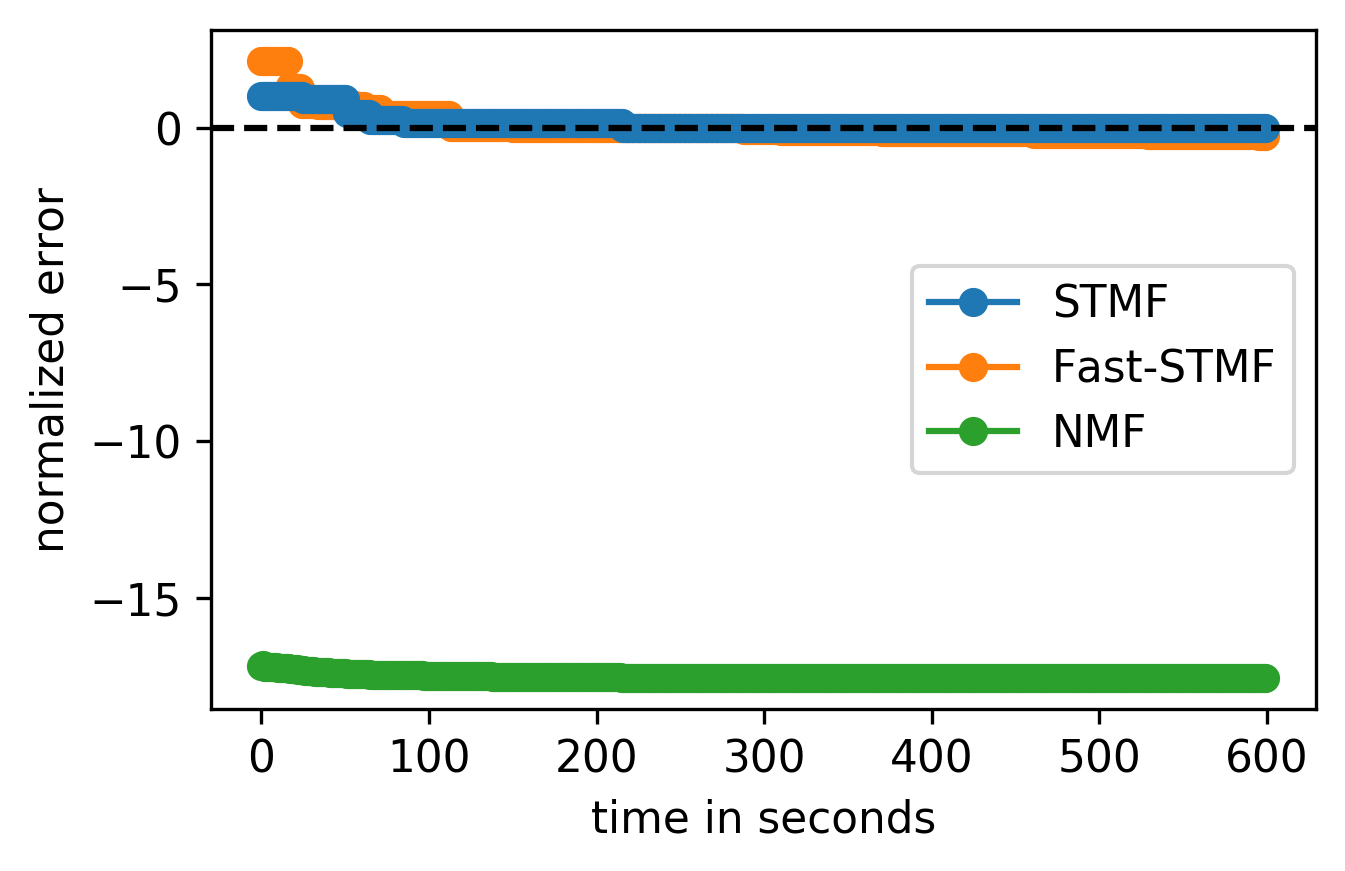}
    \caption{LUSC large}
    \label{large_lusc}
\end{figure}
\switchcolumn
\begin{figure}[H]
    \centering      
    \includegraphics[width=0.35\textwidth]{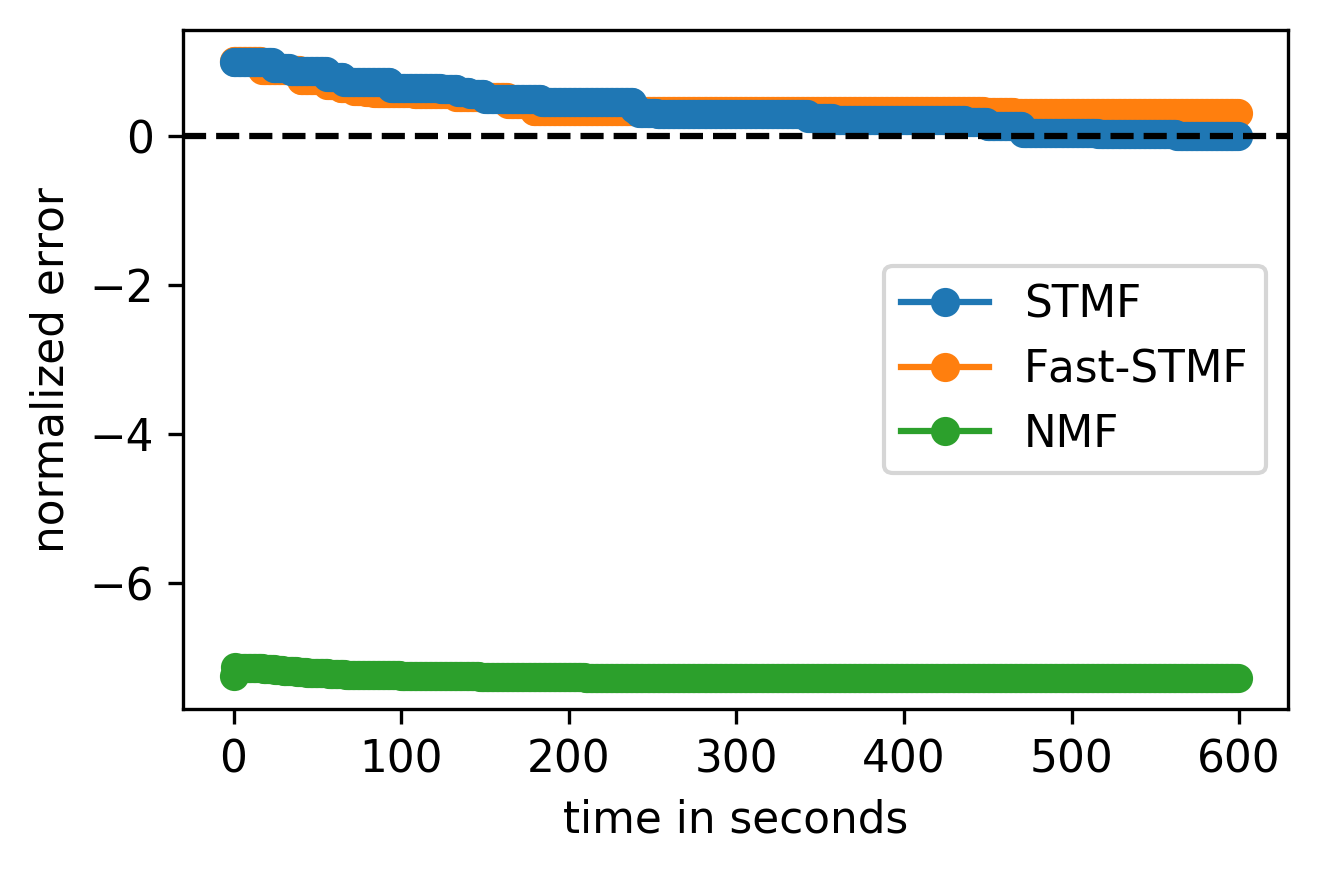}
    \caption{OV large}
    \label{large_ov}
\end{figure}
\end{paracol}

\begin{paracol}{2}
\begin{figure}[H]
    \centering
    \includegraphics[width=0.35\textwidth]{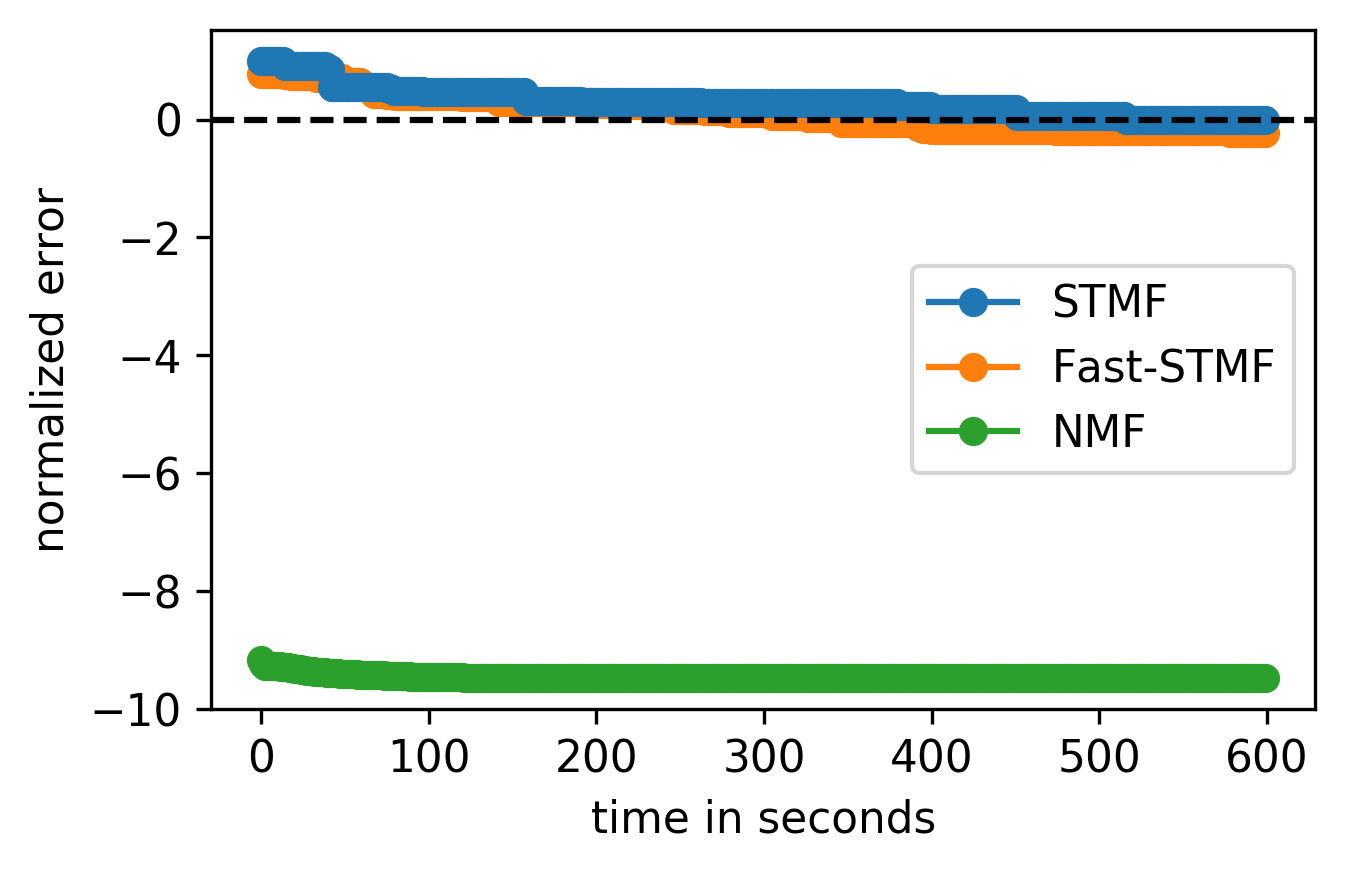}
    \caption{SARC large}
    \label{large_sarc}
\end{figure}
\switchcolumn
\begin{figure}[H]
    \centering      
    \includegraphics[width=0.35\textwidth]{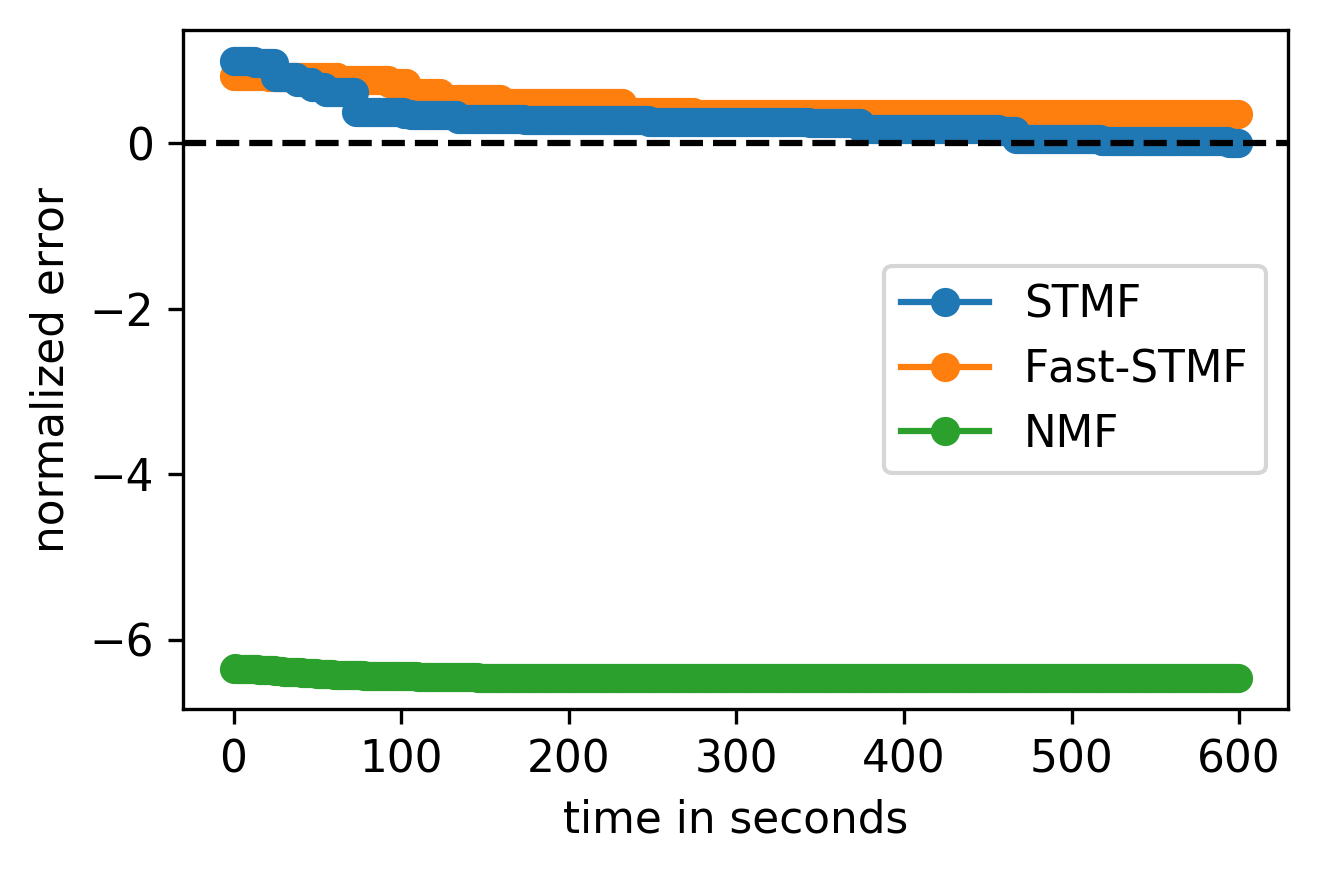}
    \caption{SKCM large}
    \label{large_skcm}
\end{figure}
\end{paracol}

\clearpage
\section{Pseudocodes}
\label{pseudocode}

% PSEUDOCODES FOR OTHER VERSIONS 
\begin{algorithm}[H]
\texttt{STMF\_ByElement\_PermC\_TD\_W} % v2
\begin{algorithmic}
\REQUIRE data matrix $R$ $\in \mathbb{R}^{m \times n}$, factorization rank $r$ 
\ENSURE factorization $U$ $\in \mathbb{R}^{m \times r}$, $V$ $\in \mathbb{R}^{r \times n}$
\STATE \textbf{if} $R$ not wide \textbf{then}  transpose $R$
\STATE $perm \gets$ permutation that orders columns in $R$ by the minimum value in the increasing order
\STATE $R \gets R[:, perm]$
\STATE initialize $U$ and compute $V$ 
        
\WHILE {not converged}
\STATE \textbf{for} each element $(i, j)$ of $R$

\begin{ALC@g}
%\STATE $k \gets td_{k}(R, U, V, i,j)$
\STATE $k \gets f(i,j)$

\STATE $(U, V, f, U_{\cdot k}',  V_{k \cdot}') \gets \text{F-ULF}(R, U, V, i, j, k)$

\STATE \textbf{if} $f$ decreases \textbf{then} \textbf{break}
\STATE \textbf{else} $(U_{\cdot k}, V_{k \cdot}) \gets (U_{\cdot k}', V_{k \cdot}')$
\end{ALC@g}
\begin{ALC@g}
\STATE $(U, V, f, U_{\cdot k}',  V_{k \cdot}') \gets \text{F-URF}(R, U, V, i, j, k)$
\end{ALC@g}
\begin{ALC@g}
\STATE \textbf{if} $f$ decreases \textbf{then} \textbf{break}
\STATE \textbf{else} $(U_{\cdot k}, V_{k \cdot}) \gets (U_{\cdot k}', V_{k \cdot}')$
\end{ALC@g}
\ENDWHILE
\STATE \textbf{if} $R$ transposed \textbf{then} $(U, V) \gets (V[:, perm^{-1}]^{T}, U^{T})$
\STATE \textbf{else} $(U, V) \gets (U, V[:, perm^{-1}])$
\RETURN $U, V$
\end{algorithmic}
\label{pseudocode_byelement}
\end{algorithm}

%%%%%%%%%%%%%%%%%%%%%%%%%%%%%%%%%%%%%%%%%%%%%%%%%%%%
\begin{algorithm}[H]
\texttt{STMF\_ByMatrix\_NoPerm\_TD\_W} % v6
\begin{algorithmic}
\REQUIRE data matrix $R$ $\in \mathbb{R}^{m \times n}$, factorization rank $r$ 
\ENSURE factorization $U$ $\in \mathbb{R}^{m \times r}$, $V$ $\in \mathbb{R}^{r \times n}$
\STATE \textbf{if} $R$ not wide \textbf{then}  transpose $R$
\STATE initialize $U$ and compute $V$ 
\WHILE {not converged}

\STATE $rows \gets \left\{\!\!\{ \td_{row}(R, U, V, i)\colon i=1,\ldots,m \right\}\!\!\}$

\STATE $cols \gets \left\{\!\!\{ \td_{col}(R, U, V, j)\colon j=1,\ldots,n \right\}\!\!\}$

\STATE \textit{errors}, $errors\_indices \gets [\,], [\,]$
\STATE \textbf{for} each \textit{element} $(i, j)$ of $R$
\begin{ALC@g}
\STATE $e \gets rows[i] + cols[j] - \td(R, U, V, i, j)$
\STATE \textbf{append} $e$ \textbf{to} \textit{errors}
%\STATE $k \gets  f(i, j)$
\STATE \textbf{append} $(i, j, f(i, j))$ \textbf{to} \textit{errors\_indices}
\end{ALC@g}
\STATE \textbf{for} each $d$ \textbf{in} argsort(\textit{errors}) in decreasing order
\begin{ALC@g}
\STATE $(i, j, k) \gets errors\_indices[d]$
\STATE  \textbf{if} $\text{random()}<0.5$ \textbf{then} $(U, V, f, U_{\cdot k}',  V_{k \cdot}') \gets \text{F-ULF}(R, U, V, i, j, k)$
\STATE \textbf{else} $(U, V, f, U_{\cdot k}',  V_{k \cdot}') \gets \text{F-URF}(R, U, V, i, j, k)$
\STATE \textbf{if} $f$ decreases \textbf{then} \textbf{break}
\STATE \textbf{else} $(U_{\cdot k}, V_{k \cdot}) \gets (U_{\cdot k}', V_{k \cdot}')$
\end{ALC@g}
\ENDWHILE
\STATE \textbf{if} $R$ transposed \textbf{then} $(U, V) \gets (V^{T}, U^{T})$
\STATE \textbf{else} $(U, V) \gets (U, V)$
%\RETURN $U[perm^{-1}, :], V$
\RETURN $U, V$
\end{algorithmic}
\label{pseudocode_bymatrix}
\end{algorithm}

\end{document}